\documentclass[runningheads]{llncs}
\usepackage{graphicx}
\usepackage{amsmath,amssymb}
\usepackage{color}
\usepackage{multirow}

\usepackage{tikz}
\usepackage{enumerate}
\usepackage{url}

\begin{document}

\pagestyle{headings}
\mainmatter
\def\ECCVSubNumber{5450}

\newcommand{\cut}[1]{\relax}

\title{Mining self-similarity: Label super-resolution with epitomic representations}


\author{Nikolay Malkin\inst{1} \and
Anthony Ortiz\inst{2} \and
Nebojsa Jojic\inst{3}}
\institute{
Yale University, New Haven, CT 06520, USA \\ \email{kolya.malkin@yale.edu} \and
Microsoft AI for Good Research Lab, Redmond, WA 98052, USA \and 
Microsoft Research, Redmond, WA 98052, USA \\ \email{\{anthony.ortiz,jojic\}@microsoft.com}
}


\maketitle

\begin{abstract}
We show that simple patch-based models, such as epitomes (Jojic et al., 2003), can have superior performance to the current state of the art in semantic segmentation and label super-resolution, which uses deep convolutional neural networks. We derive a new training algorithm for epitomes which allows, for the first time, learning from very large data sets and derive a label super-resolution algorithm as a statistical inference over epitomic representations. We illustrate our methods on land cover mapping and medical image analysis tasks.

\keywords{Label super-resolution, Semantic segmentation, Self-similarity}
\end{abstract}

\section{Introduction}

Deep convolutional neural networks (CNNs) have become a tool of choice in computer vision. They typically outperform other approaches in core tasks such as object recognition and segmentation, but suffer from several drawbacks. First, CNNs are hard to interpret, which makes them difficult to improve by adding common-sense priors or invariances into the architecture. Second, they are usually trained in a supervised fashion on large amounts of labeled data, yet in most applications labels are sparse, leading to various domain adaptation challenges. Third, there is evidence of failure of the architecture choices that were meant to promote CNNs' reasoning over large distances in images. The \emph{effective} receptive field \cite{NIPS2016_6203} of CNNs -- the distance at which faraway pixels stop contributing to the activity of deeper neurons -- is often a small fraction of the theoretical one.

With the third point in mind, we ask a simple question, the answer to which can inform an agenda in building models which are interpretable, can be pretrained in an unsupervised manner, adopt priors with ease, and are amenable to well-understood statistical inference techniques: \emph{If deep CNNs effectively use only small image patches for vision tasks, and learn from billions of pixels, then how would simple exemplar-like approaches perform, and can they be made practical computationally?} We show that models based on epitomic representations \cite{jojic2003epitomic}, illustrated in Fig.~\ref{fig:epitome_overlapping_patches}, match and surpass deep CNNs on several weakly supervised segmentation and domain transfer tasks.

\begin{figure}[t!p]
    \centering
    \begin{tabular}{ccc}
        \includegraphics[width=0.21\textwidth]{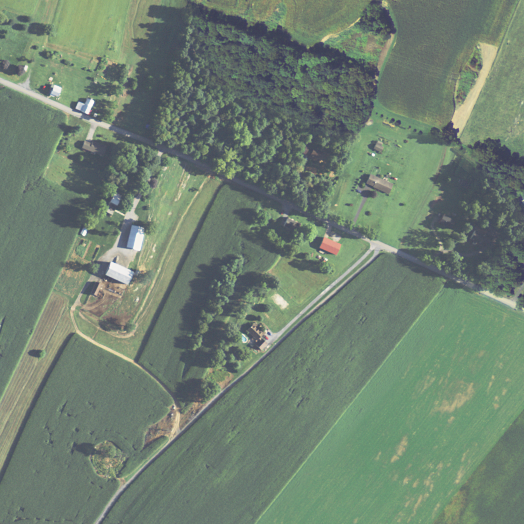}
        &
        \includegraphics[width=0.21\textwidth]{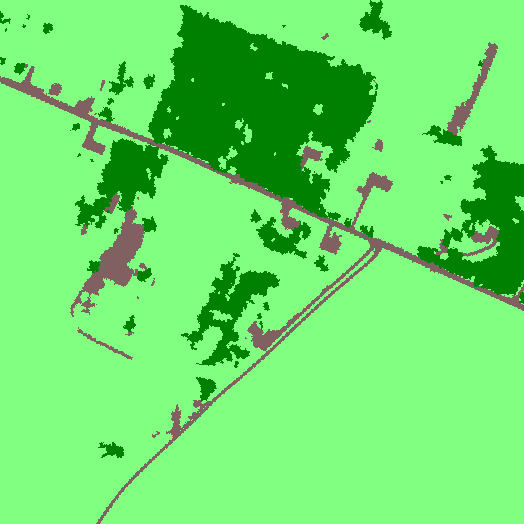}
        &
        \includegraphics[width=0.21\textwidth]{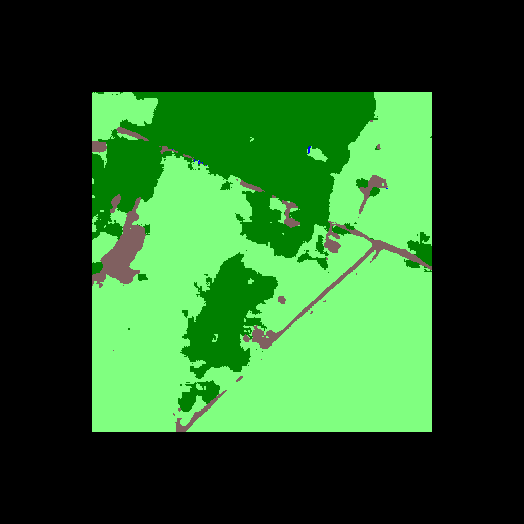}
        \\
        {\scriptsize\textit{(a)} Input}
        &
        {\scriptsize\textit{(b)} Ground truth}
        &
        {\scriptsize\textit{(c)} Prediction}
        \\
        \includegraphics[width=0.21\textwidth]{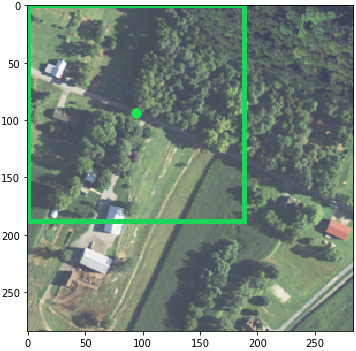}
        &
        \includegraphics[width=0.21\textwidth]{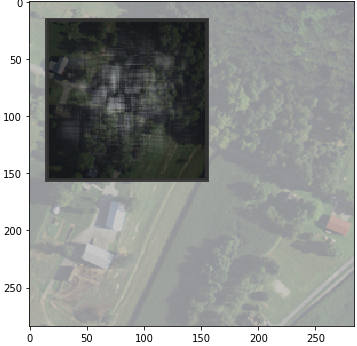}
        &
        \includegraphics[width=0.21\textwidth]{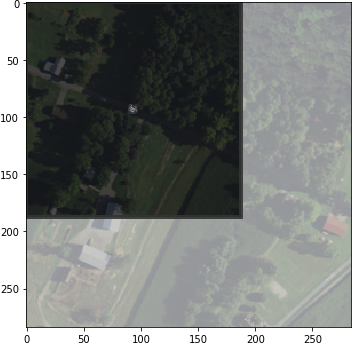}
        \\
        {\scriptsize\textit{(d)} Pixel evaluated}
        &
        {\scriptsize\textit{(e)} Grad. bottleneck}
        &
        {\scriptsize\textit{(f)} Grad. last layer}
    \end{tabular}
    \caption{Gradient-based effective receptive field estimation: We use the gradients from selected intermediate layers to the input image to estimate the size of the effective receptive field. In \textit{(e)}, we visualize the normalized gradient map (at a single coordinate shown on green in \textit{(d)}) of the U-Net's bottleneck (highest downsampling) layer with respect to the input image; \textit{(f)} shows gradients of the \emph{final} layer for the same pixel. The dark squares show the theoretical receptive field of the layers in question ($139 \times 139$ for the bottleneck and $183 \times 183$ for the final layer). However, the gradient map \textit{(f)} suggests that the effective receptive field is only about $13\times13$ pixels on average}
    \label{fig:reshuffle}
\end{figure}

For example, in Fig. \ref{fig:reshuffle} we show a patch of aerial imagery and the output of a U-Net \cite{unet} trained to predict land cover. The network misclassifies as vegetation the road pixels that appear in tree shadows. The model was trained on a large land cover map \cite{ches,uswide-cvpr} that presents many opportunities to learn that roads are long and uninterrupted. The land cover data contains many more patterns that would help see rivers through a forest, recognize houses based on their proximity to roads, etc., but the U-Nets\cut{, though having state-of-the-art performance,} do not seem to learn such long-range patterns. This myopic behavior has been observed in other architectures as well \cite{NIPS2016_6203,bagnet,texture-iclr}.

In contrast, our algorithms directly model small image patches, forgoing long-range relationships. As generative models of images, epitomes are highly interpretable: they look like the images they were trained on (Fig.~\ref{fig:epitome_overlapping_patches}). Our generative formulation of image segmentation allows the inference of labels in the latent variable space, with or without high-resolution supervision (Fig.~\ref{fig:lsr_mapping}). They achieve comparable performance to the state-of-the-art CNNs on semantic segmentation tasks, and surpass the CNNs' performance in domain transfer and weakly supervised (label super-resolution) settings.

In summary, our contributions are as follows:

\textbf{(1)} As previous training algorithms fail to fit large epitomes well, we develop new algorithms that are suitable for mining self-similarity in very large datasets.

\textbf{(2)} We develop a new label super-resolution formulation that mines image self-similarity using epitomes or directly in a single (small) image.

\textbf{(3)} We show how these models surpass the recent (neural network) state of the art in aerial and pathology image analysis.

\textbf{(4)} We illustrate that our approaches allow and even benefit from unsupervised pre-training (separation of feature learning from label embedding). 

\textbf{(5)} We show that our models deal with data size gracefully: We can train an epitome on a large fully labeled aerial imagery / land cover map and obtain better transfer in a new geography than CNNs \cite{malkin2018label,uswide-cvpr}, but we get even better results by analyzing one $512\times512$ tile at a time, with only low-resolution labels.


\begin{figure}[t!p]
    \centering
    \includegraphics[scale=0.43]{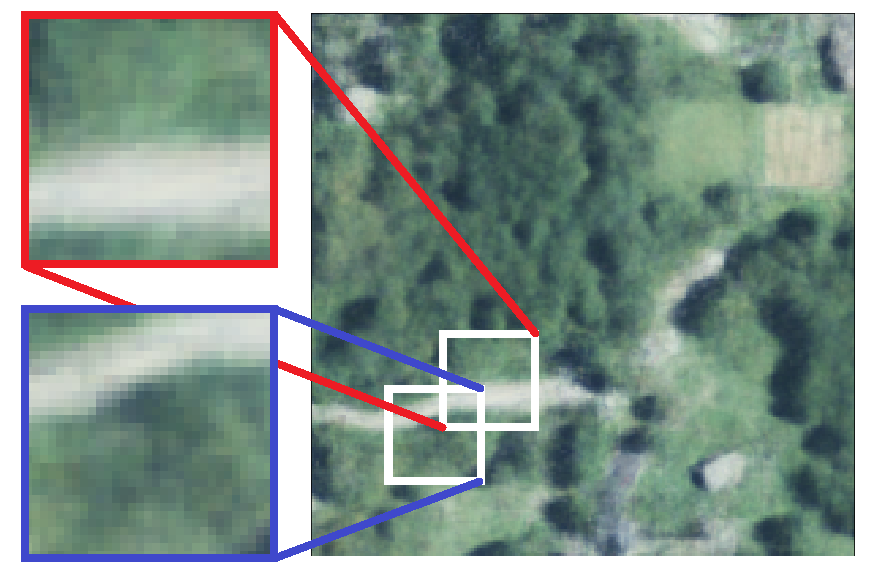}
    \includegraphics[scale=0.2]{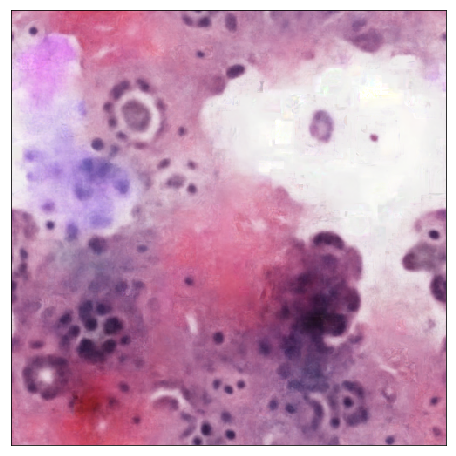}
    \caption{A quarter of an epitome ($\boldsymbol\mu$ parameters shown) trained on aerial imagery (\textit{left}) and an epitome trained on pathology slides (\textit{right}). Any $31\times31$ training data patch is generated by, and likely similar to, some $31\times31$ window in the epitome. Note the two overlapping windows: the patches are distant in color space, but their corresponding mixture components share parameters on the intersection. The epitomes are $200\times$ and $30000\times$ smaller, respectively, than their total training data}
    \label{fig:epitome_overlapping_patches}
\end{figure}

\section{Epitomes as segmentation models}

Epitomes \cite{jojic2003epitomic} are an upgraded version of a Gaussian mixture model of image patches. In this section we present, for completeness, the definition of these models. We then explain how they can be turned into segmentation models.

Consider a training set consisting of image patches $x^t$ unwrapped as vectors $x^t=\{x^t_{i,j,k}\}$, where $i,j$ are coordinates in the patch and $k$ is the spectral channel (R,G,B,\dots), and the corresponding vector of one-hot label embeddings $y_t=\{y^t_{i,j,\ell}\}$, $ \ell \in \{1,\dots,L\}$. In a mixture model, the distribution over the (image, label) data is represented with the aid of a latent variable $s\in\{1,\dots,S\}$ as
\begin{eqnarray}
    p(x^t,y^t)=\sum_{s=1}^{S} p(x^t|s)p(y^t|s)p(s),
\end{eqnarray}
where $p(s)$ is the frequency of a mixture component $s$, while the conditional probability $p(x^t|s)$ describes the allowed variation in the image patch that $s$ generates and $p(y^t|s)$ describes the likely labels for it. Under this model, the estimate for $\hat{y}$, the expected segmentation of a new image $x$, is
\begin{equation}
    p(y|x)=\sum_s p(s|x)p(y|s).
    \label{eq:pyx}
\end{equation}
A natural choice for $p(x|s)$ is a diagonal Gaussian distribution, 
\begin{equation}
    p(x|s)=\prod_{i,j,k} \frac{\exp\left({-\frac{1}{2}(x_{i,j,k}-\mu_{s,i,j,k})^2/\sigma^2_{s,i,j,k}}\right)}{(2\pi\sigma^2_{s,i,j,k})^{\frac{1}{2}}}
\end{equation}
and for $p(y|s)$ a product of categorical distributions over labels at each pixel position. The mean of the mixture component $s$ contains pixel values $\mu_{s,i,j,k}$, while the covariance matrix is expressed in terms of its diagonal elements $\sigma^2_{s,i,j,k}$, the variances of different color channels $k$ for individual pixels $i,j$.

Epitomic representations \cite{jojic2003epitomic} compress this parametrization by recognizing that patches of interest come from \emph{overlapping} regions and that different components $s$ should share parameters. The component index $s=(s_1, s_2)$ lives on a $N\times N$ grid, so $0\leq s_1,s_2\leq N-1$, and the parameters are shared:
\begin{equation}
\mu_{s,i,j,k}={\boldsymbol\mu}_{s_1+i, s_2+j,k} \qquad \sigma^2_{s,i,j,k}=\boldsymbol\sigma^2_{s_1+i,s_2+j,k}
\label{eq:epitome}
\end{equation}
(Indices are to be interpreted modulo $N$, i.e., with toroidal wrap-around.) Thus, the epitome is a large grid of parameters $\boldsymbol{\mu}_{m,n,k},\boldsymbol{\sigma}_{m,n,k}$, so that the parameters for the mixture component $s=(s_1,s_2)$ start at position $s_1,s_2$ and extend to the left and down by the size of the patch, as shown in Fig. \ref{fig:epitome_overlapping_patches}. Modeling $K\times K$ patches will take $K^2$ times fewer parameters for the similar expressiveness as a regular mixture model trained on $K\times K$ patches. The posterior $p(s|x)\propto p(x|s)p(s)$ is efficiently computed using convolutions/correlations, e.g.,
\begin{align}
    p(s_1,s_2|x)\propto \exp\sum_{i,j,k}\frac{-1}{2}\bigg(&\frac{x_{i,j,k}^2}{\boldsymbol\sigma_{s_1+i,s_2+j,k}^2}  -\frac{2x_{i,j,k}\boldsymbol\mu_{s_1+i, s_2+j,k}}{\boldsymbol \sigma_{s_1+i,s_2+j,k}^2}\nonumber\\
    &+\frac{\boldsymbol\mu_{s_1+i,s_2+j,k}^2}{\boldsymbol\sigma^2_{s_1+i,s_2+j,k}}+\log\boldsymbol\sigma_{s_1+i,s_2+j,k}^2\bigg)\cdot p(s_1,s_2).
    \label{eq:epitome_pxs}
\end{align}
Epitomes are a summary of self-similarity in the images on which they are trained. They should thus contain a much smaller number of pixels than the training imagery, but be much larger than the patches with which they are trained. Each pixel in the epitome is contained in $K^2$ patches of size $K\times K$ and can be tracked back to many different positions in many images.

Conversely, this mapping of images enables embedding of \emph{labels} into the epitome after the epitome of the images $x$ has been trained. Every location in the epitome $m,n$ will have (soft) label indicators $z_{m,n,\ell}$, computed as
\begin{equation}
    p(\ell|m,n)\propto z_{m,n,\ell}=\sum_t \sum_{{s_1,s_2}:{(m,n)\in W_{s_1,s_2}}}p(s_1,s_2|x^t)y^t_{m-s_1,n-s_2,\ell},
    \label{eq:label_embedding}
\end{equation}
where $W_{s_1,s_2}$ is the epitome window starting at $(s_1,s_2)$, i.e. the set of $K^2$ coordinates $(m,n)$ in the epitome that belong to the mixture component $(s_1,s_2)$. The posterior tells us the strength of the mapping of the patch $x^t$ to each component $s$ that overlaps the position $(m,n)$. The corresponding location in the patch of labels $y^t$ is $(m-s_1, n-s_2)$, so $y^t_{m-s_1,n-s_2,\ell}$ is added to the count $z_{m,n,l}$ of label $\ell$ at location $(m,n)$. Finally, we declare $p(y_{i,j,\ell}|s_1,s_2)\propto z_{s_1+i,s_2+j,\ell}$, allowing inference of $\ell$ for a new image patch by (\ref{eq:pyx}).

\section{A large-scale epitome training algorithm}

Epitomes have been used in recognition and segmentation tasks, e.g. \cite{russakovsky2015imagenet,nilsback2006visual,bazzani2012multiple,ni2009epitomic,papandreou2014modeling,yeung2016epitomic,papandreou2015modeling,zhang2008image}. However, the standard EM training algorithm \cite{jojic2003epitomic} that maximizes the data log-likelihood $\sum_t \log \sum_s p(x^t|s)p(s)$ is not suitable to building \emph{large} epitomes of \emph{large} data sets due to the problem of ``vanishing posterior". As training advances, the dynamic range of the posterior $p(s|x^t)$ becomes too big for machine precision, and the small probabilities are set to zero. Further parameter updates discourage mapping to these unlikely positions, leading to a die-off of chunks of ``real estate'' in the epitome. The problem is exacerbated by the size of the data (and of the epitome). Due to stability issues or computational cost, previous solutions to this \cite{stel-epitome-nips} do not allow the models to be trained on the scale on which neural networks are trained.\cut{, e.g., 200 billion pixels of aerial imagery \cite{malkin2018label,uswide-cvpr}.} 
The analogous problem exists in estimating the prior $p(s)$ over epitome positions, which also needs to have a large dynamic range. If the range is flatter (e.g., if we use a uniform prior) then maximization of likelihood requires that the epitome learn only the most frequent patterns in the data, replicating slight variations of them everywhere. As imagery is mostly uniform and smooth, this creates blurry epitomes devoid of rarer features with higher variances, like various edges and corners.

\begin{figure}[t!p]
    \centering
    \includegraphics[width=0.575\textwidth]{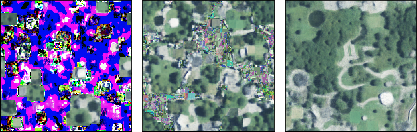}
    \hfill
    \includegraphics[width=0.38\textwidth]{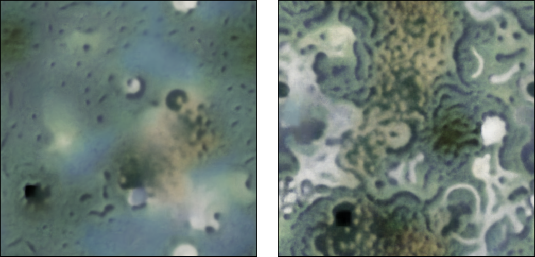}
    \caption{Numerical near-fixed points of na\"ive epitome training by SGD without location promotion, caused by vanishing posteriors, and a $399\times399$ epitome trained with location promotion (\textit{left});  non-diversifying and self-diversifying $499\times499$ epitomes trained on imagery of forests (\textit{right})}
    \label{fig:old_new_minima}
\end{figure}

Instead of EM, we develop a large-scale epitome learning algorithm combining three important ingredients: stochastic gradient descent, location promotion techniques, and the diversity-promoting optimization criterion:

\textbf{Stochastic gradient descent.} Instead of changing the parameters of the model based on all data at once, we update them incrementally in the direction of the gradient of the log-likelihood of a batch of individual data points $\frac{d}{d\theta}\log\sum_t\sum_s p(x^t|s)p(s)$, where $\theta=\{\boldsymbol\mu_{m,n,k}, \boldsymbol\sigma^2_{m,n,k}, p(s_1,s_2)\}$. 
Note that gradient descent alone does not solve the vanishing posterior problem, as the posterior also factors into the expression for the gradient (see the SI). In fact, SGD makes the situation worse (Fig. \ref{fig:old_new_minima}): the model parameters evolve before all of the data is seen, thus speeding up the extinction of the epitome's ``real estate''.

\textbf{Location promotion.} To maintain the relatively uniform evolution of all parts of the epitome, we directly constrain the learning procedure to hit all areas of the epitome through a form of posterior regularization \cite{ganchev-etal-jmlr-2010}. 
Within an SGD framework, this can be accomplished simply by keeping counters $R_{s_1,s_2}$ at each position $s_1,s_2$ and incrementing them by the posterior $p(s_1,s_2|x^t)$ upon every sample $x^t$, then disallowing mapping to the windows $s_1,s_2$ which contain the \emph{most frequently} mapped pixels. 
In particular, we compute a mask  $M=\{R_{s_1,s_2} < c/N^2\}$, where $N\times N$ is the size of the epitome, for some small constant $c<1$, and optimize only $\log\sum_{(s_1,s_2)\in M}p(x^t|s_1,s_2)p(s_1,s_2)$ at each gradient descent step. When $|M|>(1-\delta)|N|^2$ for some small $\delta$, all counters are reset to 0.

\textbf{Diversification training.} As illustrated in Fig. \ref{fig:old_new_minima} (right), standard SGD tends to learn uniform patterns, especially when trained on large datasets. Just like EM, it has to rely on the prior $p(s)$ to avoid learning blurry epitomes, but the dynamic range needed to control this is too high. Additionally, through location promotion, we in fact encourage more uniform coverage of locations. Thus, we change the optimization criterion from log-likelihood of \emph{all} data to log-likelihood of the worst modeled subset of each batch, $\sum_{t\in L_p}\sum_s p(x^t|s)$, were $L_p$ is the set of data in the worst-modeled quantile $p$ (the lowest quarter, in our experiments) in terms of data likelihood, either under a previously trained model or under the model being trained (\emph{self-diversification}). This version of a max-min criterion avoids focusing on outliers while ensuring that the data is uniformly well modeled. The resulting epitomes capture a greater variety of features, as seen in the right panel of Fig.~\ref{fig:old_new_minima}. The diversification criterion also helps the model generalize better on the test set, as we show in the experiments.


In the SI, we provide the details of the training parameters and analysis of execution time. The simple and runnable example training code\footnote{\url{https://github.com/anthonymlortiz/epitomes_lsr}} illustrates all three features of the algorithm.

\section{Label super-resolution by self-similarity}
\label{sec:lsr}
Labeling images at a pixel level is costly and time-consuming, so a number of semi-supervised approaches to segmentation have been studied, e.g., \cite{papandreou2015weakly,boxsup,hong2015decoupled,pathak2015constrained}. Recently, \cite{malkin2018label} proposed a ``label super-resolution" (LSR) technique which uses statistics of occurrence of high-resolution labels within coarse blocks of pixels labeled with a different set of low-resolution classes. (For clarity, we refer to low-res information as \emph{classes} and high-res information as \emph{labels}.) Each class, indexed by $c$, has a different composition of high-resolution labels, indexed by $\ell$.

The label super-resolution technique in \cite{malkin2018label} assumes prior knowledge of the compositions $p(\ell|c)$ of high-res labels in low-res classes and uses them to define an alternative optimization cost at the top of a core segmentation network that predicts the high-res labels. Training the network end-to-end with coarse classes results in a model capable of directly predicting the high-res labels of the individual pixels. Backpropagation through such alternative cost criteria is prone to collapse, and \cite{malkin2018label} reports best results when the data with high-res labels (HR) is mixed with data with low-res labels (LR). Furthermore, the problem is inherently ill-posed: given an expressive enough model and a perfect learning algorithm, many solutions are possible. For example, the model could learn to recognize an individual low-res block and then choose an arbitrary pattern of high-res labels within it that satisfies the counts $p(\ell|c)$. Thus the technique depends on the inductive biases of the learning algorithm and the network architecture to lead to the desirable solutions.


On the other hand, following statistical models we discuss here, we can develop a statistical LSR inference technique from first principles. The data $x$ is modeled by a mixture indexed by the latent index $s$. 
Using this index to also model the structure in the joint distribution over labels $\ell$ inside the patches generated by component $s$ and classes $c$ to which the patches belong, the known distribution of labels given the classes should satisfy $p(\ell|c)=\sum_s p(\ell|s)p(s|c)$. Thus, we find the label embedding $p(\ell|s)$ by minimizing the KL distance between the known $p(\ell|c)$ and the model's prediction $\sum_s p(\ell|s)p(s|c)$, i.e, by solving

\begin{equation}
    p(\ell|s)=\arg\max_{p(\ell|s)} \sum_{c} p(c) \sum_\ell p(\ell|c) \log \sum_s p(\ell|s)p(s|c),
    \label{eq:lsr}
\end{equation}
where $p(c)$ are the  observed proportions of low-res classes in the data and $p(s|c)$ is obtained as the posterior over $s$ for data of label $c$, as we will discuss in a moment. First, we derive an EM algorithm for solving the problem in Eq. \ref{eq:lsr} using auxiliary distributions $q_{\ell,c}(s)$ to repeatedly bound $\log \sum_s p(\ell|s)p(s|c)$ and reestimate $p(\ell|s)$. To derive the E step, we observe that
\begin{equation}
    \log \sum_s p(\ell|s)p(s|c) = \log \sum_s q_{\ell,c}(s) \frac{p(\ell|s)p(s|c)}{q_{\ell,c}(s)} \ge  \sum_s q_{\ell,c}(s) \log \frac{p(\ell|s)p(s|c)}{q_{\ell,c}(s)}. \nonumber
\end{equation}
The bound holds for all distributions $q_{\ell,c}$ and is made tight for
\begin{equation}
    q_{\ell,c}(s) \propto p(\ell|s)p(s|c).
    \label{eq:LSR_E}
\end{equation}
Optimizing for $p(\ell|s)$, we get
\begin{equation}
    p(\ell|s) \propto \sum_{c} p(c)p(\ell|c) q_{\ell,c}(s).
    \label{eq:LSR_M}
\end{equation}
Coordinate ascent on the $q_{l,c}(s)$ and $p(\ell|s)$ by iterating (\ref{eq:LSR_E}) and (\ref{eq:LSR_M}) converges to a local maximum of the optimization criterion.

Therefore, all that is needed for label super-resolution are the distributions $p(s|c)$ that tell us how often each mixture component is seen within the class $c$. Given low-res labeled data, i.e., pairs $(x^t, c^t)$ and a trained mixture model for image patches $x^t$, the answer is
\begin{equation}
    p(s|c) \propto \sum_{t:c^t=c} p(s|x^t).
    \label{eq:class_emb} 
\end{equation}
In other words, we go through all patches, look at the posterior of their assignment to prototypes $s$, and count how many times each prototype was associated with each of the classes.

\begin{figure}[t!p]
    \centering
    \includegraphics[width=0.5\textwidth]{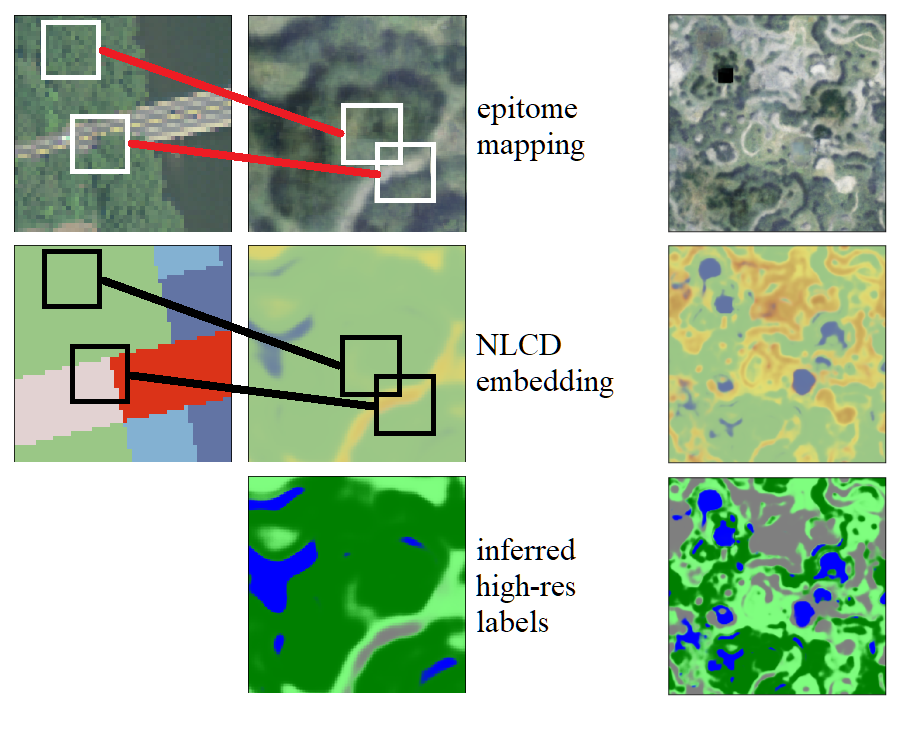}
    \caption{Two image patches are shown mapped to a piece of an epitome (\textit{left}). Below the source image, we show class labels for 30$\times$30m blocks. Below the epitome we show a piece of the class embedding $p(m,n|c)$ at a pixel level (Eqn. \ref{eq:lr_embedding_posterior}) using the same color scheme, with colors weighted by the inferred class probabilities. Below the class embedding we show the piece of the output of the label super-resolution algorithm in Section \ref{sec:lsr}. We also show the full epitome and its embeddings (\textit{right})}
    \label{fig:lsr_mapping}
\end{figure}

The epitomic representation with its parameter sharing has an additional advantage here. With standard Gaussian mixtures of patches, the level of the super-resolution we can accomplish is defined by the size of the patch $x$ we use in the analysis, because all of the reasoning is performed on the level of the patch index $s$, not at individual pixels. Thus, to get super-resolution at the level of a single pixel, our mixture model would have to be over individual pixels, i.e., a simple color clustering model (see the SI for examples). With epitomes, however, instead of using whole patch statistics, we can assign statistics $p(m,n|c)$ to individual positions in the epitome,
\begin{equation}
    p(m,n|c) \propto \sum_t \sum_{i, j} p((s_1,s_2)=(m-i,n-j)|x^t)[c^t=c],
    \label{eq:lr_embedding_posterior}
\end{equation}
where $p( \cdot, \cdot |x^t)$ is the posterior over positions. This equation represents counting how many times each \emph{pixel} in the epitome was mapped to by a patch that was inside a block of class $c$, as illustrated in Fig. \ref{fig:lsr_mapping}: While the two patches map close to each other into the epitome, the all-forest patch is unlikely to cover any piece of the road. Considering all patches in a larger spatial context, the individual pixels in the epitome can get statistics that differ from their neighbors'. This allows the inference of high-res labels $\ell$ for the entire epitome, shown with its embedding of low-res classes $c$ and super-resolved high-res labels $\ell$ on the right.

In summary, our LSR algorithm first uses the epitome model of $K \times K$ patches to embed class labels on an individual pixel level using Eq. \ref{eq:lr_embedding_posterior}. This then allows us to run the EM algorithm that iterates Eqs. \ref{eq:LSR_E} and \ref{eq:LSR_M} on positions $m,n$ associated with the shared parameters in the epitome instead of mixture components $s$, using $p(m,n|c)$ in Eq. \ref{eq:lr_embedding_posterior} in place of $p(s|c)$. Once the estimate of the high-res labels $p(\ell|m,n)$ is computed for each position in the epitome, we can predict labels in imagery using Eq. \ref{eq:pyx}. This procedure performs probabilistic reasoning over the frequencies of repeating patterns in imagery labeled with low-resolution classes to reason over individual pixels in these patterns.

\section{Experiments}
\label{sec:experiments}

\subsection{Land cover segmentation and super-resolution}
\label{sec:landcover}

\begin{figure*}[t!p]
    \centering
    \begin{tabular}{c|c|c}
    \includegraphics[width=0.3\textwidth]{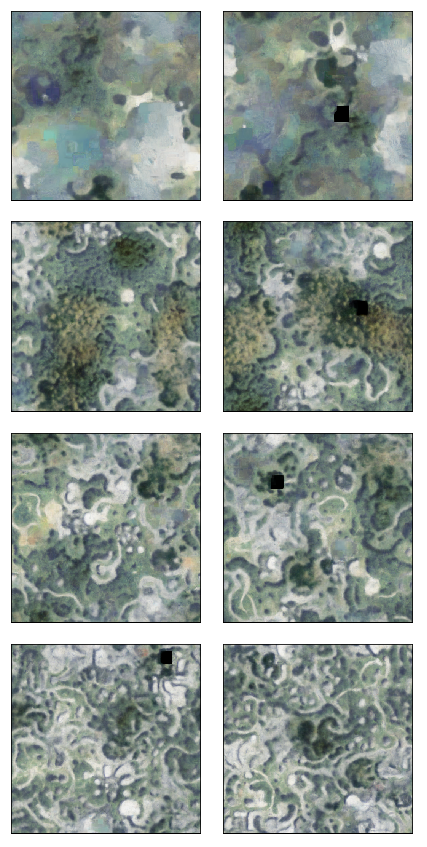}&
    \includegraphics[width=0.3\textwidth]{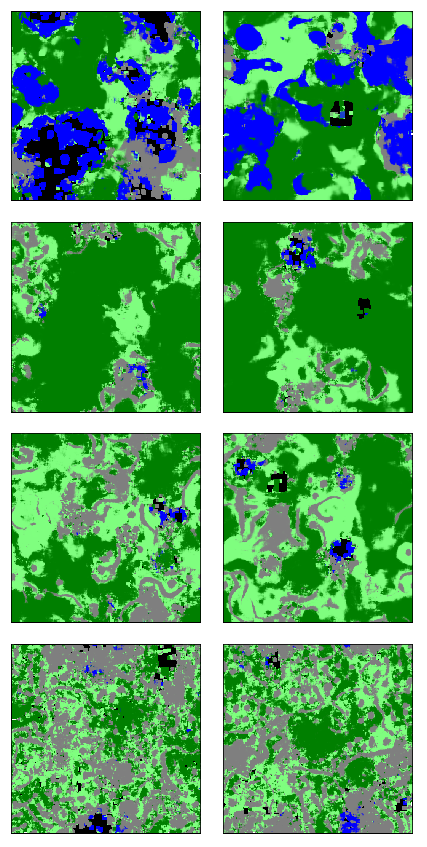}&
    \includegraphics[width=0.3\textwidth]{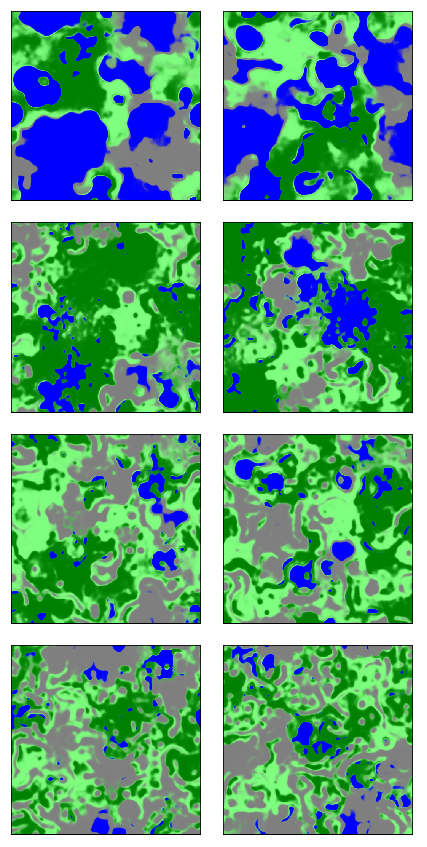}
    \end{tabular}
    \caption{Epitomes (total area $2\cdot10^6$ pixels) trained on $5\cdot10^9$ pixels of \textbf{South} imagery (\textit{left}); land cover embeddings (argmax label shown) derived from high-resolution \textbf{South} ground truth (\textit{middle}), land cover embeddings derived by epitomic LSR from \textbf{North} 30m-resolution NLCD data (\textit{right})}
    \label{fig:chesapeake_epitome}
\end{figure*}

Our first example is the problem of land cover segmentation from aerial imagery. We work with the data studied by \cite{uswide-cvpr}, available for 160,000km$^2$ of land in the Chesapeake Bay watershed (Northeast US):

    \textbf{(1)} 1m-resolution 4-band aerial imagery (NAIP) taken in the years 2013-4;
    
    \textbf{(2)} High-resolution (1m) land cover segmentation in four classes (water, forest, field/low vegetation, built/impervious) produced by \cite{ches};
    
    \textbf{(3)} Low-resolution (30m) land cover labels from the National Land Cover Database (NLCD) \cite{nlcd}. 

As in \cite{uswide-cvpr}, the data is split into \textbf{South} and \textbf{North} regions, comprising the states of MD, VA, WV, DE (S) and NY and PA (N). Our task is to produce 1m-resolution land cover maps of the \textbf{North} region, using only the imagery, possibly the low-res classes, and possibly the high-res labels from just the \textbf{South} region. The predictions are evaluated against high-res ground truth in the \textbf{North} region.

Despite the massive scale of the data, differences such as imaging conditions and frequency of occurrence of vegetation patterns make it difficult for neural networks trained to predict high-res labels from imagery in the \textbf{South} region to transfer to \textbf{North}. However, in their study of this problem using data fusion methods, \cite{uswide-cvpr} obtained a large improvement in \textbf{North} performance by multi-task training: the networks were trained to predict high-res labels with the objectives of (1) cross-entropy loss against high-res labels in \textbf{South} and (2) super-resolution loss \cite{malkin2018label} against the distributions determined by low-res NLCD labels in \textbf{North} (see the first and third rows of Table~\ref{tab:lc_sr_results}).


\textbf{Epitome training.} We fit eight $499\times499$ epitomes to all available \textbf{South} imagery. To encourage a diversity of represented land types, for each of the four high-res labels $\ell$ (water, forest, field, built), we trained a self-diversifying epitome $E_0^{(\ell)}$ on patches of size $11\times11$ to $31\times31$ containing at least one pixel labeled with label $\ell$. We then trained a model $E_1^{(\ell)}$ on the quarter of such patches with lowest likelihood under $E_0^{(\ell)}$ and a model $E_2^{(\ell)}$ on the quarter with lowest likelihood under $E_1^{(\ell)}$. The first epitome $E_0^{(\ell)}$ was then discarded.\footnote{$E_2^{(\ell)}$ is trained to model the patches poorly modeled by the self-diversifying $E_1^{(\ell)}$. Hence, $E_2^{(\ell)}$ simply has much higher posteriors and more diversity of texture.} The final model is a uniform mixture of the $E_i^{(\ell)}$ ($i=1,2$). The $\boldsymbol{\mu}_{m,n}$ parameters of its components can be seen in the left column of Fig.~\ref{fig:chesapeake_epitome}. (Notice that while the epitomes in each row were trained on patches containing pixels of a given label $\ell$, other label appear in them as well. For example, we see roads in the forest epitome (second row), since roads are sometimes found next to trees, and indeed are poorly modeled by a model of only trees, cf.\ Fig.~\ref{fig:old_new_minima}.)

\textbf{High-resolution label embedding.} We derive high-resolution soft label embeddings $p(\ell|m,n)$ from high-res \textbf{South} labels by the following procedure: for 10 million iterations, we uniformly sample a $31\times31$ patch of \textbf{South} imagery $x^t$ and associated high-res labels $y^t$ and evaluate the posterior over positions $p(s_1,s_2|x^t)$, then embed the center $11\times11$ patch of labels $y^t$ weighted by the posterior (sped up by sampling; see the SI for details). The label embeddings $p(\ell|m,n)\propto z_{m,n,\ell}$ are proportional to the sum of these embeddings over all patches; these quantities estimate the probability that a patch generated by an epitome window with center near $(s_1,s_2)$ would generate label $\ell$ at the corresponding position. These embeddings are shown in the middle column of Fig.~\ref{fig:chesapeake_epitome}.

\textbf{Low-resolution NLCD embedding.} Using the same set of epitomes trained on \textbf{South}, we derive the posteriors $p(m,n|c)$ given a low-resolution class $c$: we sample $11\times11$ patches $x^t$ from \textbf{North} with center pixel labeled with low-res class $c^t$ and embed the label $c^t$ weighted by the posterior $p(s_1,s_2|x^t)$. By (\ref{eq:lr_embedding_posterior}), $p(m,n|c)$ is then proportional to the sum of these embeddings.  An example of the embeddings in one epitome component is shown in Fig.~\ref{fig:lsr_mapping}.

\begin{figure*}[t]
    \centering
    \includegraphics[width=\textwidth]{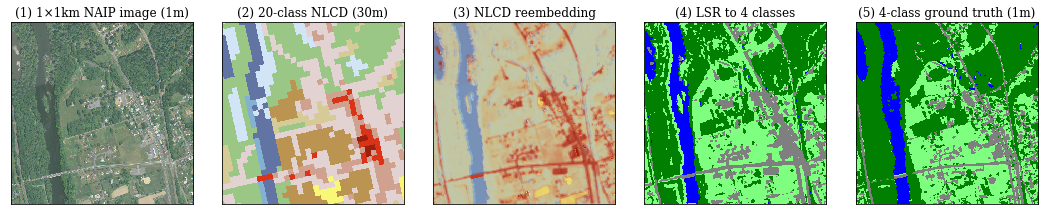}
    \caption{Self-epitomic LSR on a $1024\times1024$ patch of land (1). The low-res classes (2) are embedded at locations similar in appearance, yielding (3). The inference procedure described in Sec.~\ref{sec:lsr} produces (4), which closely resembles the ground truth (5)}
    \label{fig:lsr_big_example}
\end{figure*}

\textbf{Epitomic label super-resolution.} The joint distribution of high-res and low-res classes, $p(\ell|c)$, can be estimated on a small subset of jointly labeled data; we use the statistics reported by \cite{malkin2018label}. We apply our LSR algorithm to the low-res embeddings $p(m,n|c)$, the joint $p(c,\ell)$, and the known distribution $p(c)$ to arrive at high-res label probabilities at each epitome position, $p(\ell|m,n)$. They are shown in the right column of Fig.~\ref{fig:chesapeake_epitome}.

We evaluate the two epitome embeddings $p(\ell|m,n)$, derived from high-res labels in \textbf{South} or from low-res classes in \textbf{North}, on a sample of 1600km$^2$ of imagery in the \textbf{North} region in the following fashion: we select $31\times31$ patches $x^t$ and reconstruct the labels in the center $11\times11$ blocks as the posterior-weighted mean of the $p(\ell|m,n)$. At the large scale of data, this requires an approximation by sampling, see the SI for details. The results are shown in the second and last rows of Table~\ref{tab:lc_sr_results}.

When the area to be super-resolved is small, we can perform epitomic LSR \emph{using the imagery itself as an epitome}. We experiment with small tiles from \textbf{North} ($256\times256$ up to $2048\times2048$ pixels). For a given tile, we initialize an epitome with the same size as the tile, with uniform prior, mean equal to the true pixel intensities, and fixed variance $\sigma^2=0.01$. We then embed low-res NLCD labels from the tile into this epitome just as described above and run the LSR inference algorithm. The probabilities $p(\ell|m,n)$ are then the predicted land cover labels\footnote{We found it helpful to work with $2\times$ downsampled images and use $7\times7$ patches for embedding, with approximately $0.05|W|^2$ patches sampled for tiles of size $W\times W$.}. An example appears in Fig.~\ref{fig:lsr_big_example}, and more in the SI. The results of this \emph{self-epitomic} LSR, performed on a large evaluation set dissected into tiles of different sizes, can be seen in Table~\ref{tab:lc_sr_results}.

\begin{table}[t]
    \centering
    \caption{Performance of various methods on land cover segmentation in the \textbf{North} region. We report overall accuracy and mean intersection/union (Jaccard) index}
    \begin{tabular}{r|l|ll}
    Model&Label training set&Acc.&IoU\\\hline
         U-Net \cite{uswide-cvpr}&HR (S)\hfill&59.4\%&40.5\%\\
         Epitome (S imagery)&HR (S)\hfill&79.5&59.3 \\\hline
         U-Net neural LSR \cite{malkin2018label,uswide-cvpr}&HR (S),\hfill LR (N)&86.9&62.5\\
         \hline
         U-Net neural LSR \cite{malkin2018label}&\hfill LR (N)&80.1&41.3\\
         $256^2$ self-epitomic LSR&\hfill LR (N)&85.9&63.3\\ 
         $512^2$ self-epitomic LSR&\hfill LR (N)&87.0&65.3 \\
         $1024^2$ self-epitomic LSR&\hfill LR (N)&87.8&66.9 \\
         $2048^2$ self-epitomic LSR&\hfill LR (N)&88.0&67.8 \\
         All-tile epitomic LSR&\hfill LR (N)&83.9&58.5 \\
    \end{tabular}
    \label{tab:lc_sr_results}
\end{table}

\textbf{Results.} From Table~\ref{tab:lc_sr_results}, we draw the following conclusions:

\textit{Epitomes trained only on imagery and high-res labels in \textbf{South} transfer better to \textbf{North} than U-Nets that use the same data.} The U-Nets trained only on imagery and high-res labels in the \textbf{South} region transfer poorly to \textbf{North}: patterns associated, for example, with forests in the \textbf{North} are more frequently associated with fields in \textbf{South}, and the discriminatively trained models couple the high-frequency patterns in \textbf{South} with their associated land cover labels. Most surprisingly,  even the U-Nets trained on the LR \textbf{North} imagery perform worse than any of the epitome models trained on the same data.\footnote{We used training settings identical to those of \cite{malkin2018label}. The training collapsed to a minimum in which the ``water'' class was not predicted, but the accuracy would be lower than that of all-tile epitomic LSR even if all water were predicted correctly.}

There is evidence that the far better transfer performance of the epitomes is due to generative training. First, it is nearly unsupervised: no labels are seen in training, except to weakly guide the sampling of patches. Second, diversification training ensures, for example, that forests resembling those found in \textbf{North}, while rare, still appear in the epitomes trained on \textbf{South} imagery and receive somewhat accurate label embeddings. The posterior on those areas of the epitomes is then much higher in the \textbf{North}. (In the SI we show the mean posteriors over epitome positions illustrating this point.) 

\textit{The self-similarity in images that defines the repetition of patterns in certain classes is highly local.} If we were to study self-similarity in a large region, we would be bound to find that some imagery patterns that are associated with a particular high-res label in one area are less so in another. Therefore, the size of the area on which to perform LSR reasoning is an important design parameter. If the area is too small, then we may not get enough observations of coarse classes to unambiguously assign high-res patterns to them: indeed, self-epitomic LSR accuracy increases with the size of the tile. It is remarkable that we can get better high-res segmentation results than the state of the art by studying one $512\times512$ patch at a time, together with low-res classes for $30\times30$ blocks, and no other training data or high-res labels.

On the other hand, when the area is too large, then the pattern diversity increases and ambiguity may reduce the effectiveness of the method. Furthermore, when the area is too large, self-epitomic LSR is not computationally practicable -- the imagery must be compressed in an epitome to mine self-similarity. All-tile epitomic LSR improves over the baseline models although \emph{no high-res labels are seen}, while the best-performing U-Nets required high-res labels in \textbf{South}, low-res classes in \textbf{North}, and imagery from both \textbf{South} and \textbf{North} in training.

\subsection{Lymphocyte segmentation in pathology images}
\label{sec:pathology}

Our second example is the task of identifying tumor-infiltrating lymphocytes (TILs) in pathology imagery. We work with a set of 50000 $240\times240$ crops of 0.5$\rm\mu$m-resolution H\&E-stained tumor imagery \cite{saltz2018}. There is no high-res \emph{segmentation} data available for this task. However, \cite{hou2018} produced a set of 1786 images centered on single cells, labeled with whether the center cell is a TIL, on which our methods can be evaluated.

The best results for this task that used high-resolution supervision required either a manually tuned feature extraction pipeline and SVM classifier \cite{zhou2017,hou2018} or, in the case of CNNs, a sparse autoencoder pretraining mechanism \cite{hou2018}. More recently, \cite{malkin2018label} nearly matched the supervised CNN results using the neural label super-resolution technique: the only guidance available to the segmentation model in training was low-resolution estimates of the probability of TIL infiltration in $100\times100$ regions for the entire dataset derived by \cite{saltz2018}, as well as weak pixel-level rules (masking regions below certain thresholds of hematoxylin level).

\begin{figure}[t]
    \centering
    \includegraphics[width=0.9\textwidth]{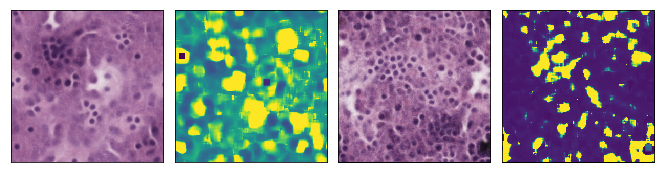}
    \caption{Epitomes trained on tumor imagery and the embedding of the tumor-infiltrating lymphocyte label. The model on the right was trained with self-diversification}
    \label{fig:lymphocyte_epitome}
\end{figure}

We address the same problem as \cite{malkin2018label}, using the low-res probability maps as the only supervision in epitomic LSR:

\textbf{Epitome training.} We train $299\times299$ epitomes on patches of size $11\times11$ to $31\times31$ intersecting the center pixels of the images to be segmented. The resulting models trained with and without self-diversification are shown in Fig.~\ref{fig:lymphocyte_epitome}.

\textbf{Low-resolution embedding.} Following \cite{malkin2018label}, we define 10 classes $c$, for each range of density estimates $[0.1\cdot n,0.1\cdot(n+1)]$. We find the posteriors $p(m,n|c)$ by embedding 1 million $11\times11$ patches from the entire dataset.

\textbf{Epitomic label super-resolution.} We estimate the mean TIL densities in each probability range, $p(\ell|c)$ and set a uniform prior $p(c)$. We then produce the probabilities of TIL presence per position $p(\ell|m,n)$ by the LSR algorithm.

We then evaluate our models on the data for which high-res labels exist by sampling $11\times11$ patches $x$ containing the center pixel -- 100 for each test image -- and computing the mean probability of TIL presence $\sum_s p(\ell|s)p(s|x)$ as the final prediction score. We obtained better results when we instead averaged the probability of TIL presence \emph{anywhere} in an embedded patch in the epitome, that is, convolved $p(\ell|s)$ with a $11\times11$ uniform filter before computing this sum.

\begin{table}[t]
    \centering
    \caption{Performance of various methods on the TIL segmentation task. We report the area under the ROC curve}
    \begin{tabular}{r|l|l}
    Model&Label training set&AUC\\\hline
    Manual features SVM \cite{zhou2017,hou2018}&HR&0.713\\
    CNN \cite{hou2018}&HR&0.494\\
    CNN with pretraining \cite{hou2018}&HR&0.786\\\hline
    U-Net neural LSR \cite{malkin2018label}&LR + color masks&0.783\\
    Non-div.\ epitomic LSR&LR&0.794 \\
    Div.\ epitomic LSR&LR&0.801 \\
    \end{tabular}
    \label{tab:lym_results}
\end{table}

\textbf{Results.} As summarized in Table~\ref{tab:lym_results}, our epitomic LSR outperforms all previous methods, including both the supervised models and the neural LSR, with self-diversifying epitomes providing the greatest improvement. The results suggest that TIL identification is a highly local problem. Deep CNNs, with their large receptive fields, require hand-engineered features or unsupervised pretraining to reach even comparable performance. In addition, epitomes are entirely unsupervised and thus amenable to adaptation to new tasks, such as classifying other types of cells: given coarse label data, we may simply embed it into the pretrained epitomes and perform LSR. 

\section{Conclusion}

Motivated by the observation that deep convolutional networks usually have a small effective receptive field, we revisit simple patch mixture models, in particular, epitomes. As generative models that allow addition of latent variables, these approaches have several advantages. They are interpretable: an epitome looks like the imagery on which it was trained (Fig.~\ref{fig:epitome_overlapping_patches}), and examining the posteriors over epitome positions is akin to understanding weights for many neurons at once. The desired invariances can be directly modeled with additional hidden variables, just as \cite{frey1999transformed} modeled illumination. They can be combined with other statistical procedures, as we show with our novel label super-resolution formulation (Sec.~\ref{sec:lsr}). They can be pretrained on a large amount of unlabeled data so that a small number of labeled points are needed to train prediction models, and they can be a base of hierarchical or pyramidal models that reason over long ranges, e.g., \cite{cheung-jojic-smaras-cvpr2007,jojic-perina-eccv2012,weber-welling-perona-2000,fergus2005learning,lazebnik-schmid-ponce-cvpr2006}. Using epitome-derived features in tasks that require long-range reasoning, such as common benchmarks for segmentation or classification of large images, is an interesting subject for future work.

Just as deep neural networks suffered from the vanishing gradient problem for years, before such innovations as stagewise pretraining \cite{stagewise}, dropout \cite{dropout}, and the recognition of the numerical advantages of ReLU units \cite{relu}, epitomic representations had suffered from their own numerical problems stemming from the large dynamic range of the posterior distributions. As a remedy, we designed a new large-scale learning algorithm that allowed us to run experiments on hundreds of billions of pixels. We showed that simply through mining patch self-similarity, epitomic representations outperform the neural state of the art in domain transfer and label super-resolution in two important application domains. 

We direct the reader to the SI for more examples, code, results on another competition dataset (in which epitomes were the basis for the winning method \cite{contest}), and discussion on future research.

\ 

\noindent\textbf{Acknowledgments:} 
The authors thank Caleb Robinson for valuable help with experiments \cite{contest} and the reviewers for comments on earlier versions of the paper.

\par\vfill\par

\clearpage

\bibliographystyle{splncs04}
\bibliography{egbib}

\begin{thebibliography}{10}
\providecommand{\url}[1]{\texttt{#1}}
\providecommand{\urlprefix}{URL }
\providecommand{\doi}[1]{https://doi.org/#1}

\bibitem{mcd12q1}
Modis/terra+aqua land cover type yearly l3 global 500m.
  \url{http://doi.org/10.5067/MODIS/mcd12q1.006}

\bibitem{ieee2019competition}
2019 ieee grss data fusion contest results.
  \url{http://www.grss-ieee.org/community/technical-committees/data-fusion/2019-ieee-grss-data-fusion-contest-results/}
  (2020)

\bibitem{codalab}
2020 ieee grss data fusion contest.
  \url{https://competitions.codalab.org/competitions/22289#results} (2020)

\bibitem{ieee2020data}
2020 ieee grss data fusion contest.
  \url{http://www.grss-ieee.org/community/technical-committees/data-fusion}
  (2020)

\bibitem{ieee2020results}
2020 ieee grss data fusion contest results.
  \url{http://www.grss-ieee.org/community/technical-committees/data-fusion/2020-ieee-grss-data-fusion-contest-results/}
  (2020)

\bibitem{bazzani2012multiple}
Bazzani, L., Cristani, M., Perina, A., Murino, V.: Multiple-shot person
  re-identification by chromatic and epitomic analyses. Pattern Recognition
  Letters  \textbf{33}(7),  898--903 (2012)

\bibitem{bagnet}
Brendel, W., Bethge, M.: Approximating {CNNs} with bag-of-local-features models
  works surprisingly well on imagenet. In: International Conference on Learning
  Representations (ICLR) (2019)

\bibitem{ches}
{Chesapeake Conservancy}: Land cover data project (January 2017),
  \url{{https://chesapeakeconservancy.org/wp-content/uploads/2017/01/LandCover101Guide.pdf}},
  [Online]

\bibitem{cheung-jojic-smaras-cvpr2007}
Cheung, V., Jojic, N., Samaras, D.: Capturing long-range correlations with
  patch models. In: 2007 IEEE Conference on Computer Vision and Pattern
  Recognition. pp.~1--8. IEEE (2007)

\bibitem{boxsup}
Dai, J., He, K., Sun, J.: Boxsup: Exploiting bounding boxes to supervise
  convolutional networks for semantic segmentation. In: Proceedings of the IEEE
  International Conference on Computer Vision. pp. 1635--1643 (2015)

\bibitem{fergus2005learning}
Fergus, R., Fei-Fei, L., Perona, P., Zisserman, A.: Learning object categories
  from google's image search. In: Tenth IEEE International Conference on
  Computer Vision (ICCV'05) Volume 1. vol.~2, pp. 1816--1823. IEEE (2005)

\bibitem{frey1999transformed}
Frey, B.J., Jojic, N.: Transformed component analysis: Joint estimation of
  spatial transformations and image components. In: Proceedings of the Seventh
  IEEE International Conference on Computer Vision. vol.~2, pp. 1190--1196.
  IEEE (1999)

\bibitem{ganchev-etal-jmlr-2010}
Ganchev, K., Gillenwater, J., Taskar, B., et~al.: Posterior regularization for
  structured latent variable models. Journal of Machine Learning Research
  \textbf{11}(Jul),  2001--2049 (2010)

\bibitem{texture-iclr}
Geirhos, R., Rubisch, P., Michaelis, C., Bethge, M., Wichmann, F.A., Brendel,
  W.: Imagenet-trained {CNNs} are biased towards texture; increasing shape bias
  improves accuracy and robustness. In: International Conference on Learning
  Representations (ICLR) (2019)

\bibitem{stagewise}
Hinton, G.E., Salakhutdinov, R.R.: Reducing the dimensionality of data with
  neural networks. science  \textbf{313}(5786),  504--507 (2006)

\bibitem{nlcd}
Homer, C., Dewitz, J., Yang, L., Jin, S., Danielson, P., Xian, G., Coulston,
  J., Herold, N., Wickham, J., Megown, K.: Completion of the 2011 national land
  cover database for the conterminous united states--representing a decade of
  land cover change information. Photogrammetric Engineering \& Remote Sensing
  \textbf{81}(5),  345--354 (2015)

\bibitem{hong2015decoupled}
Hong, S., Noh, H., Han, B.: Decoupled deep neural network for semi-supervised
  semantic segmentation. In: Advances in neural information processing systems.
  pp. 1495--1503 (2015)

\bibitem{hou2018}
Hou, L., Nguyen, V., Kanevsky, A.B., Samaras, D., Kurc, T.M., Zhao, T., Gupta,
  R.R., Gao, Y., Chen, W., Foran, D., et~al.: Sparse autoencoder for
  unsupervised nucleus detection and representation in histopathology images.
  Pattern Recognition  (2018)

\bibitem{jojic2004capturing}
Jojic, N., Caspi, Y.: Capturing image structure with probabilistic index maps.
  In: Proceedings of the 2004 IEEE Computer Society Conference on Computer
  Vision and Pattern Recognition, 2004. CVPR 2004. vol.~1, pp.~I--I. IEEE
  (2004)

\bibitem{jojic2003epitomic}
Jojic, N., Frey, B.J., Kannan, A.: Epitomic analysis of appearance and shape.
  In: ICCV. vol.~3, p.~34 (2003)

\bibitem{stel-epitome-nips}
Jojic, N., Perina, A., Murino, V.: Structural epitome: a way to summarize
  one’s visual experience. In: Advances in neural information processing
  systems. pp. 1027--1035 (2010)

\bibitem{adam}
Kingma, D., Ba, J.: Adam: A method for stochastic optimization. International
  Conference on Learning Representations  (12 2014)

\bibitem{lazebnik-schmid-ponce-cvpr2006}
Lazebnik, S., Schmid, C., Ponce, J.: Beyond bags of features: Spatial pyramid
  matching for recognizing natural scene categories. In: 2006 IEEE Computer
  Society Conference on Computer Vision and Pattern Recognition (CVPR'06).
  vol.~2, pp. 2169--2178. IEEE (2006)

\bibitem{NIPS2016_6203}
Luo, W., Li, Y., Urtasun, R., Zemel, R.: Understanding the effective receptive
  field in deep convolutional neural networks. In: Lee, D.D., Sugiyama, M.,
  Luxburg, U.V., Guyon, I., Garnett, R. (eds.) Advances in Neural Information
  Processing Systems 29, pp. 4898--4906. Curran Associates, Inc. (2016),
  \url{http://papers.nips.cc/paper/6203-understanding-the-effective-receptive-field-in-deep-convolutional-neural-networks.pdf}

\bibitem{malkin2018label}
Malkin, K., Robinson, C., Hou, L., Soobitsky, R., Czawlytko, J., Samaras, D.,
  Saltz, J., Joppa, L., Jojic, N.: Label super-resolution networks.
  International Conference on Learning Representations  (2019)

\bibitem{relu}
Nair, V., Hinton, G.E.: Rectified linear units improve restricted boltzmann
  machines. In: Proceedings of the 27th international conference on machine
  learning (ICML-10). pp. 807--814 (2010)

\bibitem{ni2009epitomic}
Ni, K., Kannan, A., Criminisi, A., Winn, J.: Epitomic location recognition.
  IEEE transactions on pattern analysis and machine intelligence
  \textbf{31}(12),  2158--2167 (2009)

\bibitem{nilsback2006visual}
Nilsback, M.E., Zisserman, A.: A visual vocabulary for flower classification.
  In: 2006 IEEE Computer Society Conference on Computer Vision and Pattern
  Recognition (CVPR'06). vol.~2, pp. 1447--1454. IEEE (2006)

\bibitem{papandreou2015weakly}
Papandreou, G., Chen, L.C., Murphy, K.P., Yuille, A.L.: Weakly-and
  semi-supervised learning of a deep convolutional network for semantic image
  segmentation. In: Proceedings of the IEEE international conference on
  computer vision. pp. 1742--1750 (2015)

\bibitem{papandreou2014modeling}
Papandreou, G., Chen, L.C., Yuille, A.L.: Modeling image patches with a generic
  dictionary of mini-epitomes. In: Proceedings of the IEEE Conference on
  Computer Vision and Pattern Recognition. pp. 2051--2058 (2014)

\bibitem{papandreou2015modeling}
Papandreou, G., Kokkinos, I., Savalle, P.A.: Modeling local and global
  deformations in deep learning: Epitomic convolution, multiple instance
  learning, and sliding window detection. In: Proceedings of the IEEE
  Conference on Computer Vision and Pattern Recognition. pp. 390--399 (2015)

\bibitem{pytorch}
Paszke, A., Gross, S., Chintala, S., Chanan, G., Yang, E., DeVito, Z., Lin, Z.,
  Desmaison, A., Antiga, L., Lerer, A.: Automatic differentiation in {PyTorch}.
  In: NIPS Autodiff Workshop (2017)

\bibitem{pathak2015constrained}
Pathak, D., Krahenbuhl, P., Darrell, T.: Constrained convolutional neural
  networks for weakly supervised segmentation. In: Proceedings of the IEEE
  international conference on computer vision. pp. 1796--1804 (2015)

\bibitem{jojic-perina-eccv2012}
Perina, A., Jojic, N.: Spring lattice counting grids: Scene recognition using
  deformable positional constraints. In: European Conference on Computer
  Vision. pp. 837--851. Springer (2012)

\bibitem{uswide-cvpr}
Robinson, C., Hou, L., Malkin, K., Soobitsky, R., Czawlytko, J., Dilkina, B.,
  Jojic, N.: Large scale high-resolution land cover mapping with
  multi-resolution data. In: Proceedings of the IEEE Conference on Computer
  Vision and Pattern Recognition. pp. 12726--12735 (2019)

\bibitem{contest}
Robinson, C., Malkin, K., Hu, L., Dilkina, B., Jojic, N.: {Weakly supervised
  semantic segmentation in the 2020 IEEE GRSS Data Fusion Contest}. Proceedings
  of the International Geoscience and Remote Sensing Symposium  (2020)

\bibitem{unet}
Ronneberger, O., Fischer, P., Brox, T.: U-net: Convolutional networks for
  biomedical image segmentation. In: International Conference on Medical image
  computing and computer-assisted intervention. pp. 234--241. Springer (2015)

\bibitem{russakovsky2015imagenet}
Russakovsky, O., Deng, J., Su, H., Krause, J., Satheesh, S., Ma, S., Huang, Z.,
  Karpathy, A., Khosla, A., Bernstein, M., et~al.: Imagenet large scale visual
  recognition challenge. International journal of computer vision
  \textbf{115}(3),  211--252 (2015)

\bibitem{saltz2018}
Saltz, J., Gupta, R., Hou, L., Kurc, T., Singh, P., Nguyen, V., Samaras, D.,
  Shroyer, K.R., Zhao, T., Batiste, R., et~al.: Spatial organization and
  molecular correlation of tumor-infiltrating lymphocytes using deep learning
  on pathology images. Cell reports  \textbf{23}(1), ~181 (2018)

\bibitem{schmitt2020sen12ms}
Schmitt, M., Hughes, L.H., Qiu, C., Zhu, X.X.: Sen12ms--a curated dataset of
  georeferenced multi-spectral sentinel-1/2 imagery for deep learning and data
  fusion. ISPRS Journal of Photogrammetry and Remote Sensing  (2020)

\bibitem{schmitt2020weakly}
Schmitt, M., Prexl, J., Ebel, P., Liebel, L., Zhu, X.X.: Weakly supervised
  semantic segmentation of satellite images for land cover mapping--challenges
  and opportunities. arXiv preprint arXiv:2002.08254  (2020)

\bibitem{dropout}
Srivastava, N., Hinton, G., Krizhevsky, A., Sutskever, I., Salakhutdinov, R.:
  Dropout: a simple way to prevent neural networks from overfitting. The
  journal of machine learning research  \textbf{15}(1),  1929--1958 (2014)

\bibitem{weber-welling-perona-2000}
Weber, M., Welling, M., Perona, P.: Unsupervised learning of models for
  recognition. In: European conference on computer vision. pp. 18--32. Springer
  (2000)

\bibitem{yeung2016epitomic}
Yeung, S., Kannan, A., Dauphin, Y., Fei-Fei, L.: Epitomic variational
  autoencoders  (2016)

\bibitem{zhang2008image}
Zhang, H., Fritts, J.E., Goldman, S.A.: Image segmentation evaluation: A survey
  of unsupervised methods. computer vision and image understanding
  \textbf{110}(2),  260--280 (2008)

\bibitem{zhou2017}
Zhou, N., Yu, X., Zhao, T., Wen, S., Wang, F., Zhu, W., Kurc, T., Tannenbaum,
  A., Saltz, J., Gao, Y.: Evaluation of nucleus segmentation in digital
  pathology images through large scale image synthesis. In: Medical Imaging
  2017: Digital Pathology. vol. 10140, p. 101400K. International Society for
  Optics and Photonics (2017)

\end{thebibliography}

\appendix

\section{Code}

\begin{figure*}
\centering
\includegraphics[width=0.8\textwidth]{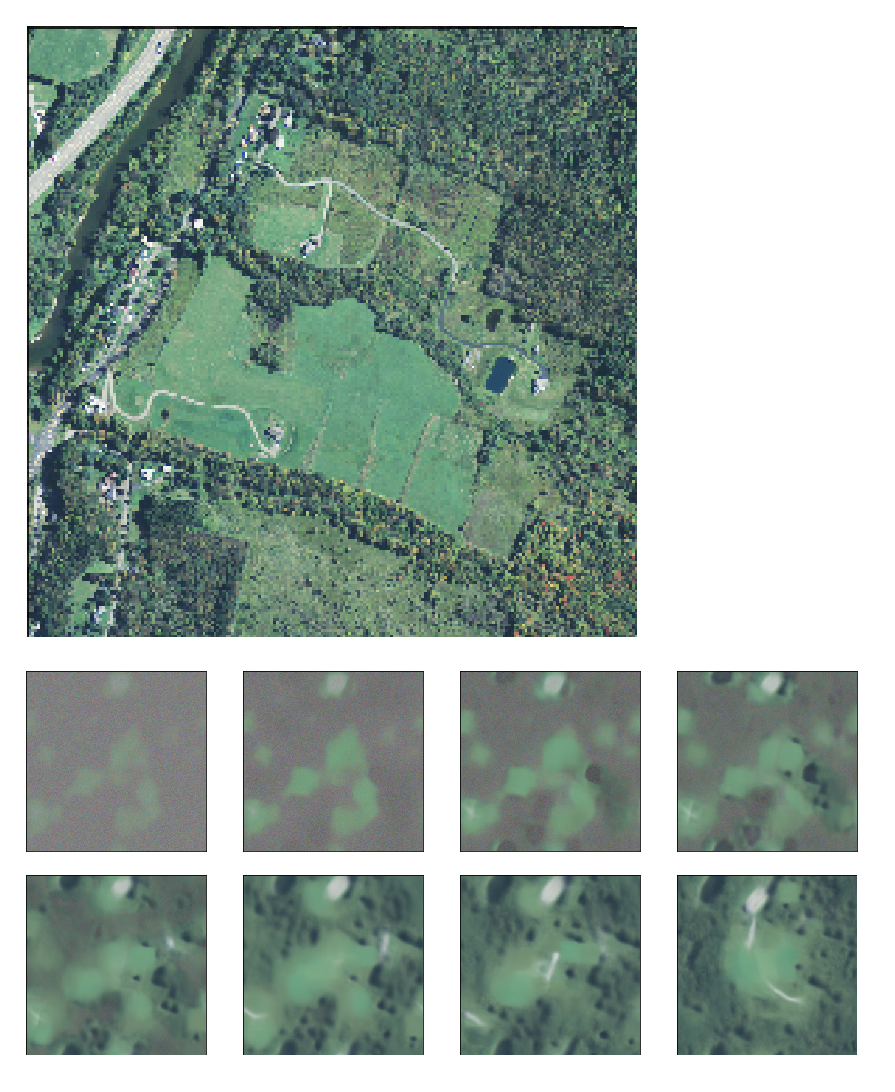}
\caption{The $1024\times1024$ image from our sample code (\textit{above}) and $299\times299$ epitome $\boldsymbol\mu$ parameters at selected iterations of epitome training (after $32, 64, 128, \dots, 4096$ batches of 256 patches), shown at the same scale (\textit{below}) }
\label{fig:epitome_iterations}
\end{figure*}

We provide sample code at \url{https://github.com/anthonymlortiz/epitomes_lsr}. The folder \texttt{data/} includes an example of a $2\times$ downsampled $2\times2$km tile of aerial imagery and the associated high-res and low-res labels. Run the notebook \texttt{train\_epitome.ipynb} to visualize the iterations of epitome training on this area. Run \texttt{self\_epitomic\_sr.ipynb} to visualize single-image label super-resolution. (The corresponding static HTML files can also be used for viewing. Some outputs of the first notebook are shown in Fig.~\ref{fig:epitome_iterations}.) 

The training code is short and clear, and we feel that it can serve in lieu of pseudocode. The reader can immediately run it, perhaps even with their own data.

\section{The vanishing posterior problem in EM and SGD epitome learning}

As derived in \cite{jojic2003epitomic}, the original EM algorithm for epitome learning optimizes the log likelihood 
\begin{equation}
    \sum_t \log \sum_{s=1}^{S} p(x^t|s)p(s)
    \label{eq:ll}
\end{equation}
by iterating two steps. First, the image patches $x^t$  are mapped to the epitome using the posterior $p(s|x) \propto p(x|s)p(s)$, with 
\begin{align}
\log p(s_1,s_2|x)={\rm const} + \sum_{i,j,k}\frac{-1}{2}\bigg(&\frac{x_{i,j,k}^2}{\boldsymbol\sigma_{s_1+i,s_2+j,k}^2}  -\frac{2x_{i,j,k}\boldsymbol\mu_{s_1+i, s_2+j,k}}{\boldsymbol \sigma_{s_1+i,s_2+j,k}^2}\nonumber\\
    &+\frac{\boldsymbol\mu_{s_1+i,s_2+j,k}^2}{\boldsymbol\sigma^2_{s_1+i,s_2+j,k}}+\log\boldsymbol\sigma_{s_1+i,s_2+j,k}^2\bigg)+\log p(s_1,s_2).
    \label{eq:app_epitome_pxs}
\end{align}
Thus, the E step is efficiently performed with convolutions, creating the posterior $p(s|x)$ used in the M step which re-estimates epitome parameters by averaging the pixels from the patches based on their locations and posterior mapping. For example, the mean epitome is re-estimated as
\begin{equation}
    \mu_{m,n,k}=\frac{\sum_t \sum_{(s_1,s_2)} x^t_{m-s_1, n-s_2,k} p(s_1,s_2|x^t)}  {\sum_t \sum_{s_1,s_2}  p(s_1,s_2|x^t)}
    \label{eq:mu_update}
\end{equation}
where inner summations over $s_1,s_2$ in numerator and denominator are performed over windows $W_{s_1,s_2}$ containing the pixel $m,n$: for a $K \times K$ patch $x^t$, $(s_1,s_2)$ goes from $(m-K+1, n-K+1)$ to $(m,n)$. We readily recognize that both numerator and denominator consist of convolutions of the $K \times K$ patch $x^t$ with the posterior $N\times N$ posterior map $p(s_1,s_2|x^t)$ over an $N \times N$ epitome.

The main numerical difficulty with epitome training is that the posterior map can be zero in many places, as the consequence of a large dynamic range of the values computed in Eq. \ref{eq:epitome_pxs}, especially as the variances contract during learning. Upon exponentiation to obtain $p(s|x)$, some locations in the epitome may get their probability clipped to zero. If all data patches avoid certain pixels in the epitome, then those locations stop being updated. Note that this is not a local minimum problem, but purely a numerical precision issue. Avoiding division by zero by adding a small constant to the numerator and the denominator in Eq. \ref{eq:mu_update} and subtracting the maximum in Eq.~\ref{eq:epitome_pxs} before exponentiation do nothing to remedy this problem: The areas of the epitome that die off simply get filled with the mean of all pixels seen in training. In practice EM training usually has to be carefully coaxed out of these situations, for example by training first with larger patches and later with smaller ones. The larger patches hit more pixels, making it less likely that some epitome locations will never be part of the component $s$ for which at least one data point $x^t$ shows a measurable posterior after exponentiation. For a regular mixture model the problem can be remedied by keeping track of the highest value of the posterior $p(s|x^t)$ over all data points. In epitomes, because of parameter sharing, this is not trivial. In \cite{stel-epitome-nips}, the authors introduce a solution which evaluates different epitome parts with different levels of precision. The number of precision levels is finite, and the complexity of the learning algorithm scales linearly with the number of levels, which makes the approach impractical for very large epitomes where we could have very large dynamic ranges in the posterior.

The numerical difficulties arising from vanishing posteriors persist if the model is trained by stochastic gradient descent on the parameters of the model. For a mixture model $p(x)=\sum_sp(x|s)p(s)$, with the mixture components $p(x|s)=p(x;\theta_s)$ and the prior  $p(s)\propto e^{\pi_s}$, the posterior factors into the gradient of the log-likelihood:
\begin{align*}
    \frac{d\log p(x;\theta,\pi)}{d\theta_{s_1,s_2}}&=p(s_1,s_2|x;\theta,\pi)\frac{d\log p(x;\theta)}{d\theta_{s_1,s_2}},\\
    \frac{d\log p(x;\theta,\pi)}{d\pi_{s_1,s_2}}&=p(s_1,s_2|x;\theta,\pi).
\end{align*}
In the situation of epitomes, some $\theta_s$ are shared among different $s$. The gradients with respect to parameters at a given position vanish if all posteriors in a window around the position vanish. Furthermore, batch training and the posterior sampling described below can make it even more likely to loose parts of the epitome during training. As described in Section 3 of the main text, we use a location promotion strategy to solve this problem.

\section{Large scale epitome training details}

\subsection{Training details and parameters}

All of our epitome models were implemented using the PyTorch package \cite{pytorch} for efficient computation on the GPU; we also have efficient Matlab implementations for CPU. The epitomes are trained on variable-size patches, $11\times11$ to $31\times31$. For patches of size $w$, we smooth the probabilities $p(x^t|s)$ by temperature $(w/11)^2$. The means are initialized randomly, distributed as $0.5+\text{unif}(0,0.1)$. For stability, the variances are parametrized by their inverses $1/\sigma^2$, initialized at 10 and clipped between 1 and 100, and the priors are parametrized in the log domain, initialized at 0 and clipped between $-4$ and 4. They were trained for 50000 iterations with a batch size of 256 (aerial imagery) or 30000 iterations with a batch size of 64 (pathology imagery), using the Adam optimizer \cite{adam} with initial learning rate 0.003.

For the location promotion mechanism in training, we use a threshold of $10^{-8}\cdot\text{(batch size)}$. For the $499\times499$ aerial imagery epitomes, this amounts to $c\approx0.64$, and for the $299\times299$ pathology epitomes, $c\approx0.057$. The counter reset threshold is $\delta=0.05$: after $0.95$ of the epitome has hit its threshold, the counters are zeroed.

\subsection{Evaluation by sampling}

Here we explain the high-res label embedding and reconstruction method mentioned in Section 5.1. For a given patch $x$ with high-res labels $y$, we compute the posterior distribution of epitome positions $p(s|x)$. To embed the labels $y$ -- that is, to add them, appropriately weighted, to the the counts that give $p(\ell|m,n)$, it would be necessary to convolve the $11\times11$ patch of \emph{labels} with the posterior. It is more efficient, and equivalent in expectation, to sample several locations from the posterior distribution and to add the labels $y$ to the counts surrounding those locations. We use 16 samples in our experiments. (Because the posterior distributions tend to be peaky, this does not affect the results greatly.) This also explains why there are unmapped areas in the middle column of Fig.~5: the eight epitomes were trained individually, but when they were combined as components of a uniform mixture, those areas were captured by the other epitomes and were never sampled. Similarly, in reconstruction, for a given patch $x$, we sample several positions $s^*_n$ from the posterior distribution and sum the labels in windows around the $s^*_n$ to form the output predictions for $x$.

\subsection{Execution time}

In this section we report the execution times for both U-Nets and epitome algorithms used in our land cover experiments. The EM algorithm for label super-resolution, which consists of iterative matrix multiplications and normalizations, runs in not more than a few seconds, and the main computational cost is incurred in training and evaluation.

The training of one set of four land cover epitomes and of a U-Net both take about 12 hours in our implementation. Note that the two models were implemented in different neural net toolkits: the U-Net implementation in the CNTK toolkit by the authors of \cite{malkin2018label,uswide-cvpr} is two years old and ran on a different version of the CUDA library than our implementation of epitomes in Torch.

The average evaluation times per pixel with different methods are reported in Table~\ref{tab:execution}. For the epitomes, the execution time does not include the label embedding (which can be done incrementally at training time) or the (fast) label super-resolution inference algorithm. On the other hand, for the self-epitomic LSR results, we sampled 5\% of all $7\times7$ patches in the input tile for the low-resolution label embedding; this computation forms the bulk of the computation time, since the LSR algorithm immediately produces the target high-resolution labels without the need to reevaluate the posterior mappings.

We see a number of optimizations to the epitome algorithms that are likely to speed up the evaluation. For example, some terms in the expression for the posterior are independent of the input patch and can be precomputed. In addition, our epitome code does not use the optimized multithreaded data loading methods that appear in the U-Net implementation.

\begin{table}[]
    \caption{Execution times (per pixel) of various algorithms on the land cover data.}
    \centering
    \begin{tabular}{r|c|c}
        Method & parameters & time / thousand pixels \\ \hline
         U-Net & $256\times256$ input & 3ms \\  \hline
         All-tile epitomic LSR & $11\times11$ windows, one sample & 53ms \\
         All-tile epitomic LSR & $31\times31$ windows, one sample & 29ms \\
         Self-epitomic LSR & $128\times128$ input & 7ms \\
         Self-epitomic LSR & $256\times256$ input & 8ms \\
         Self-epitomic LSR & $512\times512$ input & 25ms \\
         Self-epitomic LSR & $1024\times1024$ input & 94ms  \\
    \end{tabular}
    \label{tab:execution}
\end{table}

\section{On domain transfer: Comparison of the posterior distributions for land cover regions}

\begin{figure*}
    \centering
    \begin{tabular}{c|c|c}
    \includegraphics[width=0.25\textwidth]{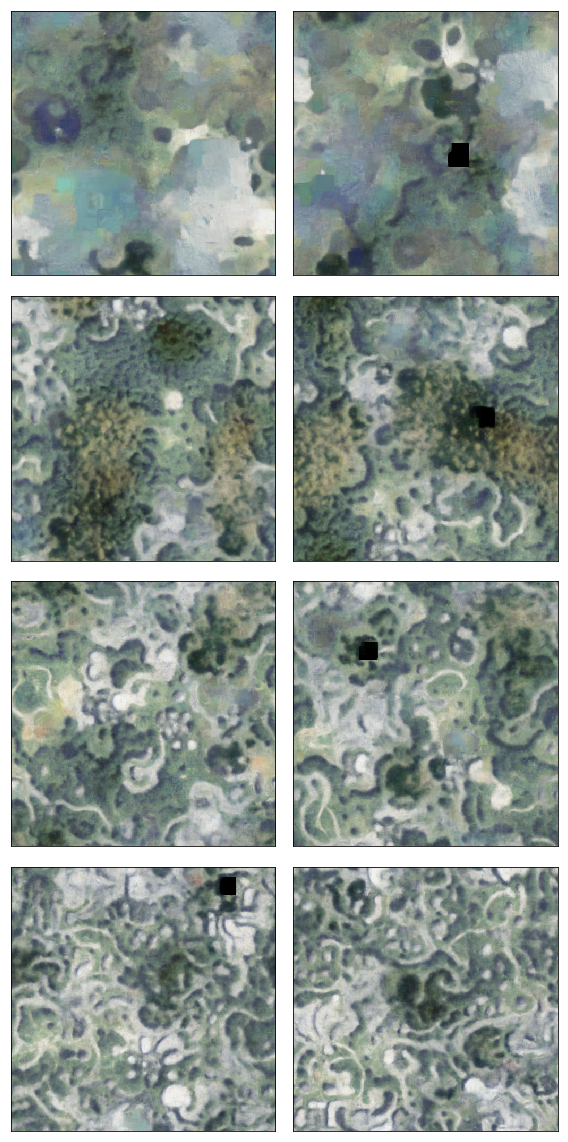}&
    \includegraphics[width=0.25\textwidth]{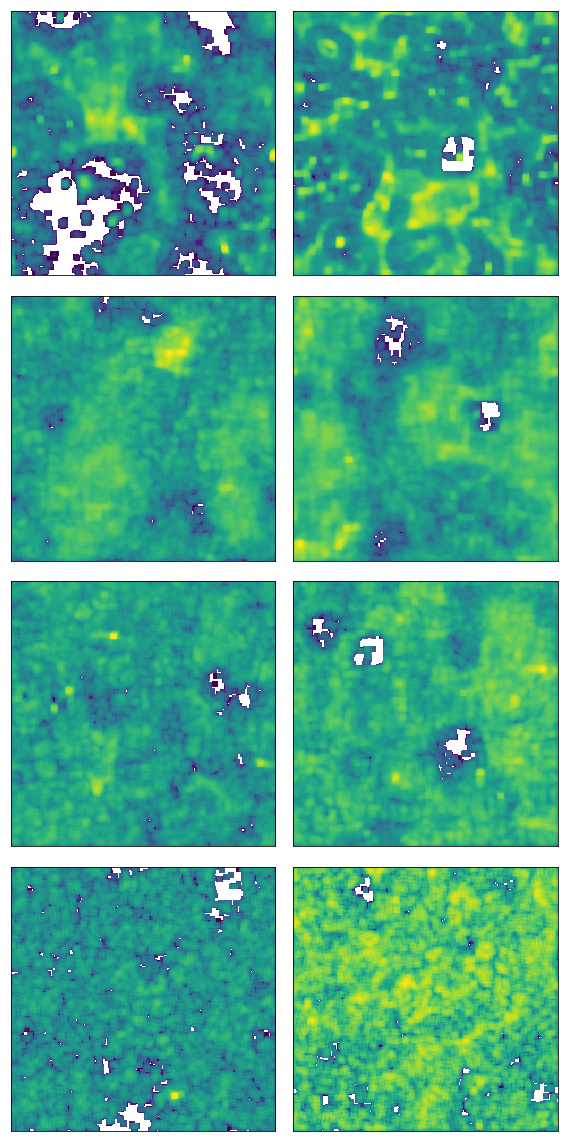}&
    \includegraphics[width=0.25\textwidth]{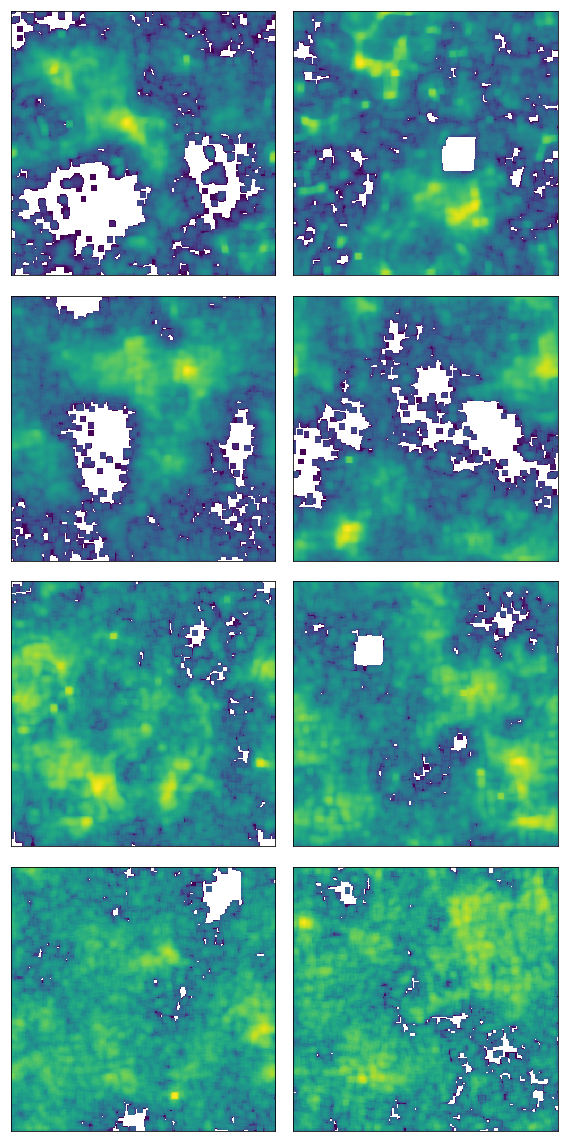}
    \end{tabular}
    \caption{Epitomes (total area $2\cdot10^6$ pixels) trained on imagery from the \textbf{South} region (\textit{left}), the log posterior over positions for $31\times31$ patches in the \textbf{South} region (\textit{middle}), the log posterior over positions for $31\times31$ patches in the \textbf{North} region (\textit{right})}
    \label{fig:posteriors}
\end{figure*}

Fig.~\ref{fig:posteriors} shows the eight components of the land cover epitome (Fig.~5 in the main text) and the posterior distributions in the \textbf{South} and \textbf{North} regions. Observe that the \textbf{South} data is more uniformly mapped. Also, notice the different distribution in forested areas: for example, the light-green forests are rare in \textbf{North}. The change in data distribution is in this way easily detectable and explain why U-Nets do not transfer from the \textbf{South} to the \textbf{North}. Our epitome training is based on the diversification criterion targeting only the distribution over the input patches $x^t$, but favoring the worst modeled patches. The forests of the \textbf{North} where U-Nets make errors can actually be found in the \text{South}, too, but they are just more rare. The diversification training learns these patterns, too, and the epitomic classification is not confused by these patterns in the \text{North}, resulting in 20\% increase in accuracy (see the top of Table 1 in the main text).

\section{Label super-resolution}

\label{sec:app_lsr}

In Fig. \ref{fig:nlcd_legend} we provide the standard National Land Cover Database (NLCD) class descriptions \cite{nlcd} and color legend in Fig. \ref{fig:nlcd_legend}. We use the same colors in all NLCD visualizations here and in the main text. In Table \ref{tab:nlcd_stats} we also provide the $p(\ell|c)$ statistics from \cite{malkin2018label} which we used in our experiments.

In Fig. \ref{fig:LSRexamples} we show a few more results of the single-tile LSR technique for visual inspection of both accurate and inaccurate predictions. But first, as promised in the main text, we address the possibility of performing LSR simply using pixel colors.

As discussed in the main text, our label super-resolution technique iterates equations
\begin{equation}
    q_{\ell,c}(s) \propto p(\ell|s)p(s|c).
    \label{eq:app_LSR_E}
\end{equation}
and
\begin{equation}
    p(\ell|s) \propto \sum_{c} p(c)p(\ell|c) q_{\ell,c}(s).
    \label{eq:app_LSR_M}
\end{equation}
Here, we assume that the texture components indexed by $s$ are associated with classes $c$ through $p(s|c)$, and the iterative procedure finds association between the components and high-res labels $\ell$, i.e. the distribution $p(\ell|s)$. A straightforward interpretation of this is that components $s$ are associated with image patches, and so $p(\ell|s)$ only tells us the (probability of) label $\ell$ of the whole prototype $s$, and therefore the patches mapped to it. Thus, without the epitomic reasoning in Eq.~\ref{eq:LSR_E} in the main text, the granularity with which we can assign high-res labels would depend on the patch size, and to get to the highest possible resolution, we would have to assume that $s$ indexes the prototypes of a \emph{single} pixel size. In other words, $s$ would simply refer to a color clustering of the image (precisely, a component of a mixture model on colors).

\begin{figure}[t]
\centering
\includegraphics[width=0.99\textwidth]{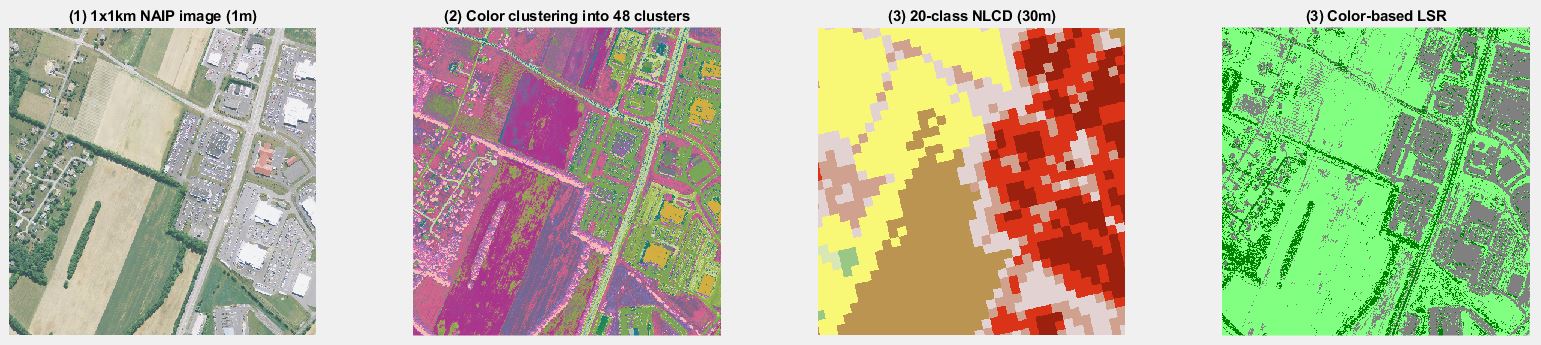}
\caption{Color LSR}
\label{fig:colorLSR}
\end{figure}

Fig. \ref{fig:colorLSR} illustrates such an approach. Using a single 1km$^2$ RGBI tile, we train a mixture of 48 4-dimensional diagonal Gaussians indexed by $s$ and show the cluster assignments for each pixel to these 48 clusters using a random color palette. The term $p(s|c)$ is then computed simply by computing the $48 \times 20$ matrix of co-occurrences of clusters $s$ and coarse classes $c$, which are also shown in the figure, and then normalizing the matrix appropriately. Iterating the two equations above now leads to the $4\times 48$ mapping of labels $\ell$ to components $s$ in form of the distribution $p(\ell|s)$. When the most likely label is assigned to each cluster, we obtain the LSR result shown in the last panel in the figure. The resulting HR predictions are remarkable given that only pixel colors were used,  but they also, unsurprisingly, exhibit a scattering of predicted impervious pixels all over the image, and in general a fairly speckled result. In addition, the model confuses the main road in the patch with fields and forests. On the other hand, the epitomic pixel-wise reasoning described in the main text yields the result shown in the first row of Fig. \ref{fig:LSRexamples}, where the road is accurately predicted and the speckles are generally suppressed, indicating that our technique reasons about larger patterns and the individual pixels within them to assign single-pixel-level labels.

\begin{figure}[p]
    \centering
       { \includegraphics[width=0.8\textwidth]{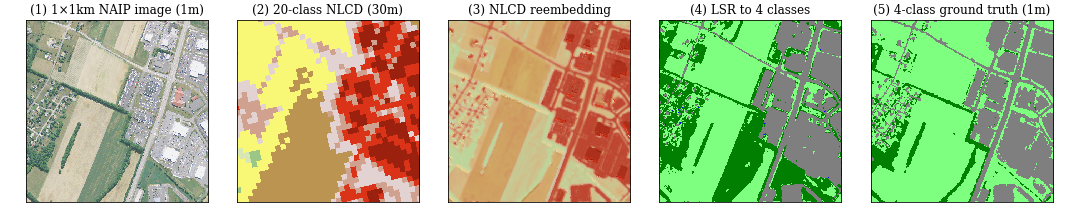} }
    \\
       { \includegraphics[width=0.8\textwidth]{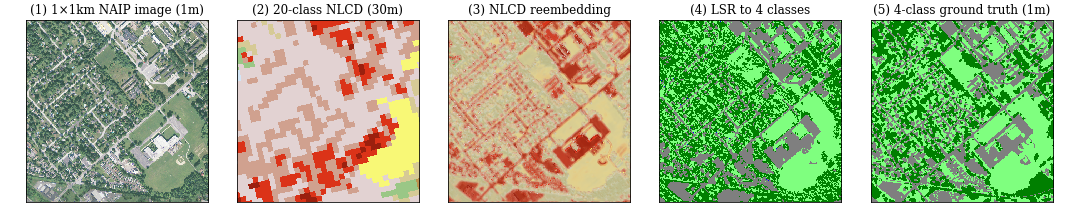} }
    \\
       { \includegraphics[width=0.8\textwidth]{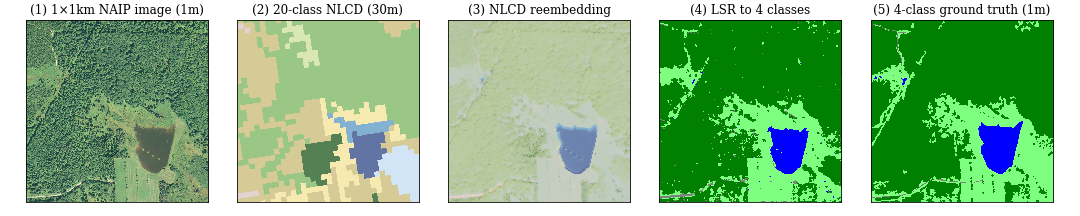} }
    \\
       { \includegraphics[width=0.8\textwidth]{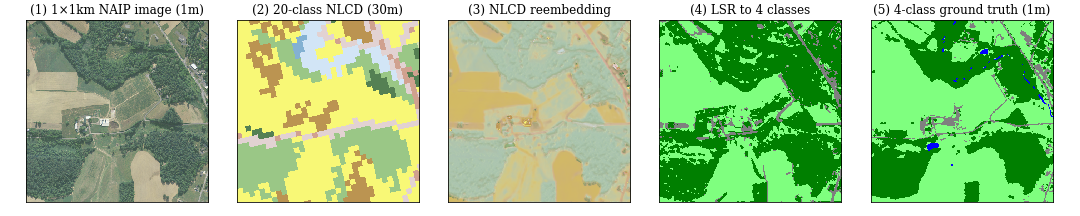} }
    \\
       { \includegraphics[width=0.8\textwidth]{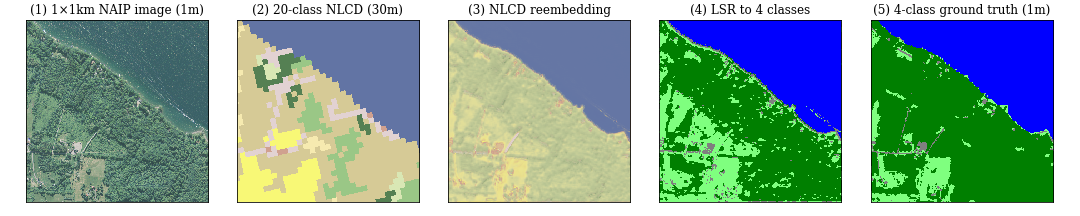} }
    \\
       { \includegraphics[width=0.8\textwidth]{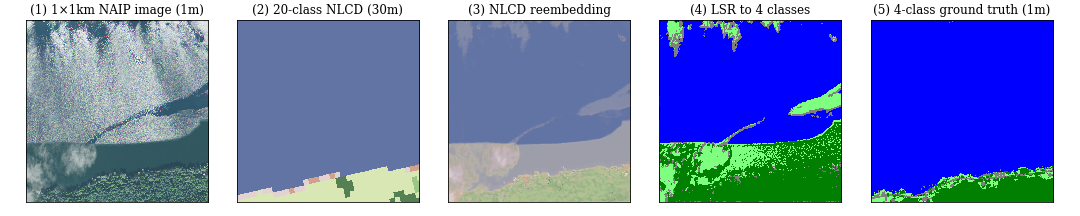} }
    \\
       { \includegraphics[width=0.8\textwidth]{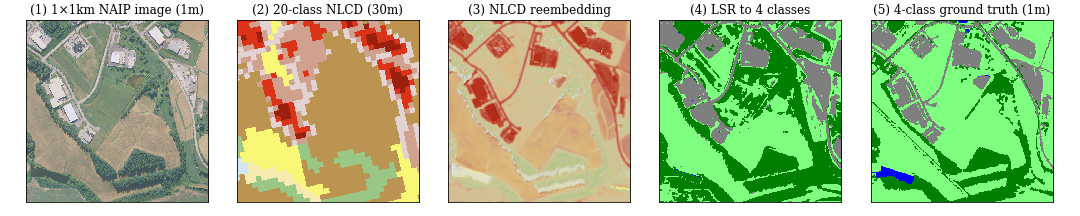} }
    \caption{Examples of single-tile (self-epitomic) LSR using $7\times7$ patches on $2\times$ downsampled images. The first row provides the analysis of the same tile shown super-resolved with a simple color model in Fig.~\ref{fig:colorLSR}. The rest are selected to illustrate both the success and failure modes of the single-tile approach}
    \label{fig:LSRexamples}
\end{figure}

\begin{table}[h]
\caption{Statistics $p(\ell|c)$ of the four HR labels in 15 of the 20 land cover classes that appear in our data. Statistics are from \cite{malkin2018label}. Our model did not use the available uncertainty in the statistics, but priors can easily be added in our model (see the future work section)}
\centering
\begin{tabular}{r|c|c|c|c}
{NLCD class}&{water}&{forest}&{field}&{imperv.}\\ \hline
Open water&.97&.01&.01&.02\\ \hline
Developed, Open Space&.00&.42&.46&.11\\
Developed, Low &.01&.31&.34&.35\\
Developed, Medium &.01&.14&.21&.63\\
Developed, High &.01&.03&.07&.89\\ \hline
Barren Land &.09&.13&.45&.32\\ \hline
Deciduous Forest&.00&.92&.06&.01\\
Evergreen Forest&.00&.94&.05&.01\\
Mixed Forest&.01&.92&.06&.02\\ \hline
Shrub/Scrub&.00&.71&.26&.03\\
Grassland/Herbaceous&.01&.38&.54&.07\\
Pasture/Hay&.00&.11&.86&.03\\
Cultivated Crops&.00&.11&.86&.03\\ \hline
Woody Wetlands&.01&.90&.08&.00\\
Emergent Wetlands&.11&.07&.81&.01 \\
\end{tabular}
\label{tab:nlcd_stats}
\end{table}

\begin{figure}[p]
    \centering
    \includegraphics[width=0.49\textwidth]{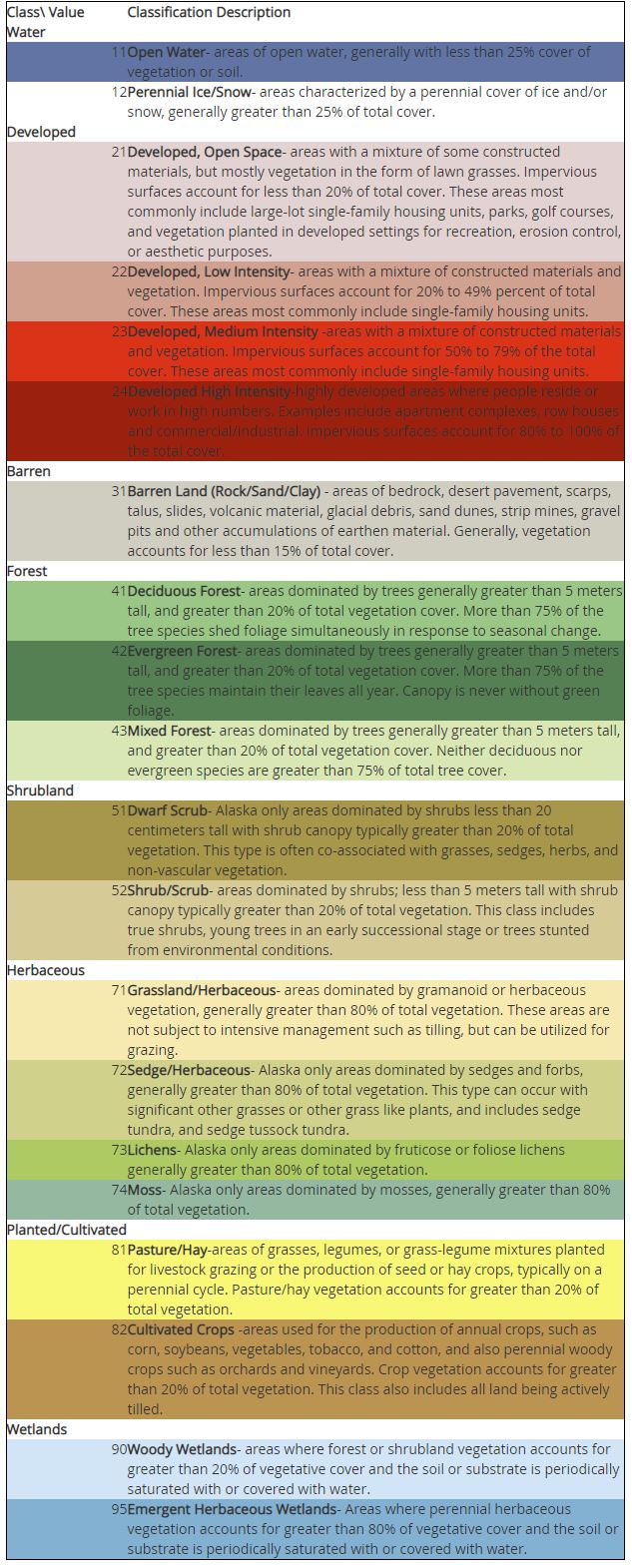}
    \caption{National land cover database (NLCD) legend: descriptions and the color code \cite{nlcd}}
    \label{fig:nlcd_legend}
\end{figure}

\section{Epitomic LSR in the 2020 IEEE-GRSS Data Fusion Contest}

The IEEE Geoscience and Remote Sensing Society ran a competition on a task almost identical to the label super-resolution application on land cover described in the main paper. We show here that we can reach the top result in the competition through analysis \emph{only} of the validation set (986 images), without ever looking at the very large training set of over 180k pairs of satellite images and their low-res labels. A detailed description of the winning method can be found in \cite{contest}.

The goal of the 2020 IEEE GRSS data fusion competition \cite{codalab} was to infer high-resolution (10m) labels in 8 classes (forest, shrubland, grassland, wetland, cropland, urban, barren, water) based on  high-resolution (10m) 12-band Sentinel imagery and low-resolution (500m) MODIS labels (Fig.~\ref{fig:sentinel-modis-examples}). To that end, a training set of around 180k $256\times256$ Sentinel images \cite{schmitt2020sen12ms} and the corresponding MODIS labels \cite{mcd12q1} was provided, and the competition was performed on class average accuracy on a validation set (986 images) and a test set. We examine the validation set stage of the competition, in which 70 teams officially participated. The ground truth high-resolution labels for the validation set have since been made publicly available \cite{ieee2020data}, making the following analysis possible.

The baseline methods \cite{schmitt2020weakly} used both random forests and deep CNNs, which all achieved average accuracy not more than $54.1$\%. The top 10 teams' average accuracies on the validation set ranged between 68\% and 71\%. While we do not know how much data these teams used and which methods were tested by the teams in the competition, given the number of participants, it is probably safe to assume that convolutional neural networks and random forests were used in a variety of ways, as they were by the other winning teams in both this \cite{ieee2020results} and last year's \cite{ieee2019competition} contests. Yet, our approach is a straightforward mixture model as described here and in the main text.

\begin{figure}[h]
    \centering
    \includegraphics[width=0.8\textwidth]{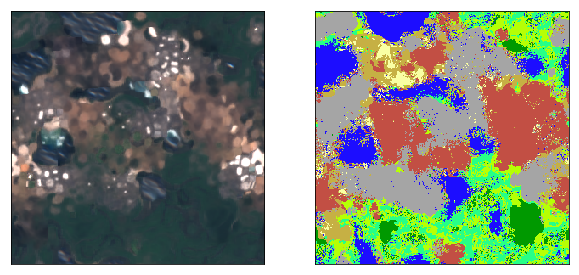}
    \caption{$299\times299$ epitome of Sentinel imagery (RGB channels shown) and the output of the LSR algorithm, coded using the standard 10-class color scheme from \cite{schmitt2020weakly}}
    \label{fig:sentinel_epitome}
\end{figure}

\begin{figure}[h]
    \centering
    \includegraphics[width=0.8\textwidth]{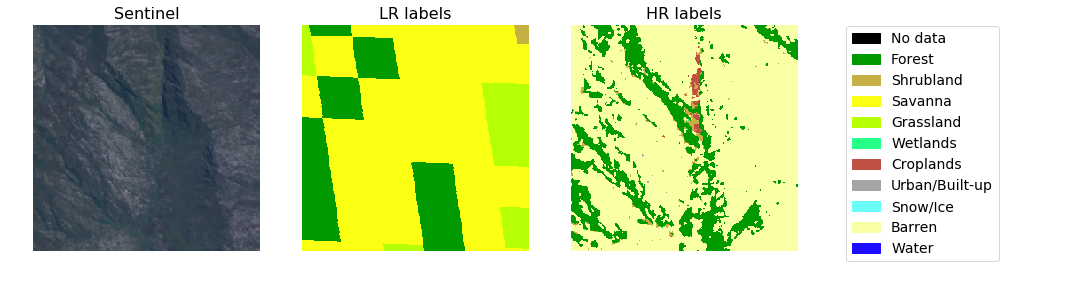}\\
    \includegraphics[width=0.8\textwidth]{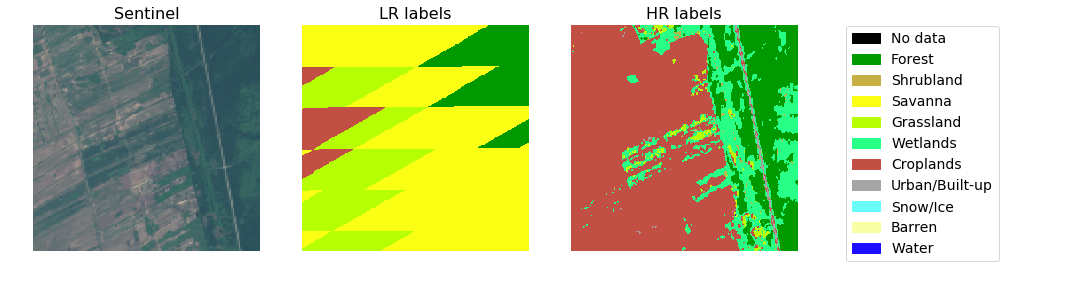}\\
    \includegraphics[width=0.8\textwidth]{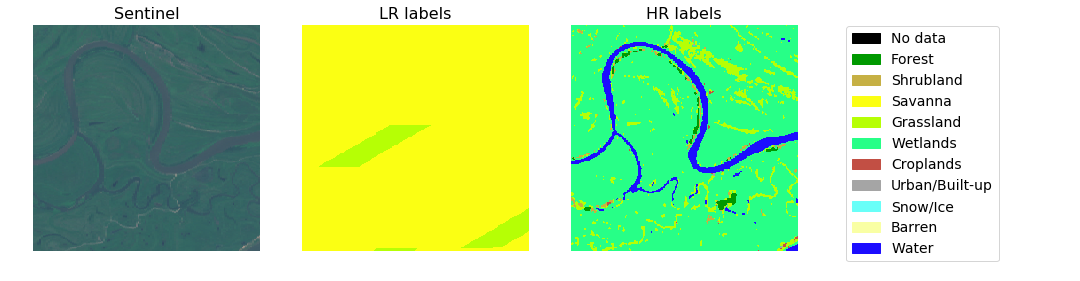}\\
    \caption{Examples of high-res imagery, low-res labels, and high-res labels from the IEEE GRSS competition dataset}
    \label{fig:sentinel-modis-examples}
\end{figure}

In particular, we start with a simple color clustering model similar to the one in Sec.~\ref{sec:lsr}, equivalent to learning an epitome with patch size $1\times1$, because Sentinel imagery has many more bands than NAIP imagery, making separation of certain classes easier. For example, richer spectral information simplifies separating water from urban/built surfaces, as well as both of those from vegetation\footnote{This color clustering algorithm is augmented with a \emph{bag of clusters} latent variable that makes the model sensitive to nearby colors. That approach is mostly independent of this work, so we direct the reader to \cite{contest} for the details.}. However, classifying the rest of the classes (forests, grassland, shrubland, barren, and croplands) requires analyzing spatial patterns, so we built an $299\times299$ epitome model with $7\times7$ patches (Fig.~\ref{fig:sentinel_epitome}). LSR is performed in each model, using as $p(\ell|c)$ the statistical table provided by Fig.~3  in \cite{schmitt2020weakly}. Then, we ensemble these predictions.

The color-only model ($1\times1$ patches) achieves an average accuracy of $65.4$\%. The epitome model ($7\times7$ patches) yields $65.5$\%. Either score would rank as 14/70. We ensemble these by trusting the color-only model's predictions on the easily confused \emph{shrubland} and \emph{barren} classes to get average accuracy $68\%$, placing the model at 11/70. 

However, it is important to note that the ground truth labels themselves are not entirely accurate, as they were created in a semi-manual manner, rather than fully manually \cite{schmitt2020weakly}. \cut{In fact, we found that they disagree with the human labels on 3000 randomly selected points in the validation imagery.
This means that the average accuracies as well as the team rankings have to be taken with a grain of salt. In fact, using the confusion matrix on these 3000 points and randomization tests, we estimate that a model performing at 65\% accuracy is not statistically significantly worse than the top model in terms of expectation on the performance on \emph{real} human labels. } The artifacts in the ground truth labels (Fig.~\ref{fig:sentinel-modis-examples}) suggest that this `ground truth' is actually based on a convolutional network's predictions (probably on lots of HR data). 

To verify this, we trained a small neural network (a 5-layer fully convolutional network with two $3\times3$ convolutions and three $1\times1$ convolutions) on the predictions of our best ensemble model to introduce similar inductive biases, and this network reaches 70.7\%, the top score on the leaderboard. Similar models, applied to the test set, also reached the highest accuracy of 57.5\% in the main competition track.

In summary, using the simple statistical models described in this paper, it is possible to match the top score in an international competition in weakly supervised land cover mapping. Furthermore, our unsupervised learning approach is much more data-efficient than most modern-day supervised learning techniques, as evidenced by the fact that our results needed only 986 out of over 180k images in training.

\section{Future work}

We and other researchers interested in the epitomic representations discussed here have many possible follow-up research directions to explore. 

As our models are easy to interpret, they can be used in scenarios where the predictions or just their errors have to be explained. The errors of the model as a predictor of high-res labels can be tracked not only to individual patches in the epitome, but through the epitome back to other patches in the data. In fact, as we discussed above, the visualization in Fig. \ref{fig:posteriors} can be used to understand the domain shift of other models, such as U-Nets.

Traceability back to the training patches can also be used to further improve models through efficient hierarchical matching where the input patch is ultimately compared with an individual training patch, rather than its compressed version in the epitome.

It is also interesting to think about encoding various types of invariances as latent variables in the model, e.g. global illumination or local deformation variables.

The results in Fig. \ref{fig:LSRexamples} all use the same statistics in Table \ref{tab:nlcd_stats} in inference, even though, clearly, the actual statistics of occurrence of different land cover labels in different NLCD classes will vary from image to image, indicating that the model is fairly robust to errors in these statistics. It would be interesting to see just how robust the approach is, and in particular, if the simplified summary statistics described in Fig. \ref{fig:nlcd_legend} could be used, or if it is possible to super-resolve a user's estimate of label percentage in a user-defined area in human-in-the-loop schemes. 

While our label super-resolution formulation only uses the average frequencies of labels $\ell$ in different classes $c$ shown in Table \ref{tab:nlcd_stats} and applies these on the entire dataset, or an entire single tile, the recent LSR work \cite{malkin2018label} also used the uncertainty estimates (variances on the frequencies across different 30m blocks). In their approach,  a training cost is defined so that it uses both means and variances to match the statistics over predicted HR labels in each block, rather than summarily across the entire image. It is possible to use this information in our method, too. E.g., the label frequency variance can be used to set an appropriate Dirichlet prior on the $p(\ell|c)$ for each tile or even an individual 30m block, and then treat the parameters $p(\ell|c)$ as latent. 

To infer labels for an image using the epitomes and $p(\ell|s)$ maps in Fig.~5 of the main text, we need to sample a large number of patches, covering the entire image, and compare them with the epitomes. In our current implementation, the inference time is within an order of magnitude of the U-Net's computational cost. However, the inference can be sped up through experimentation with sampling strategies. It would also be interesting to study approximate, e.g., coarse-to-fine, epitome mapping techniques, or even learnable indexing techniques that would speed up the label inference.

It would be interesting to test epitomic representations on recognition tasks.

Previous work on epitomic representations recognized that patch models are powerful in modeling local patterns larger than the patch size, as the patches effectively get quilted into larger patterns, and just a few longer range relationships can go a long way towards grounding the quilting or capturing relationships useful in recognition
\cite{cheung-jojic-smaras-cvpr2007,jojic-perina-eccv2012}. Through a similar type of reasoning in models based on large scale epitomes it may be possible to further improve segmentation and recognition performance, e.g., in case of long features such as rivers and roads.

Recent work such as \cite{bagnet} has shown that the current state of the art in recognition is highly dependent on texture features, rather than image shapes. But, models we describe here can also be built on mask (shape) patches, as was shown in the paper that introduced epitomes \cite{jojic2003epitomic}, and also used in early co-segmentation work \cite{jojic2004capturing}. 

Beyond inspiration from those epitome works, we can use a variety of generative modeling work (before GANs and autoencoders), as discussed in the conclusions of the main paper, to build hierarchical models. But, using neural models of texture in combination with statistical reasoning is also possible.

\end{document}